\documentclass{clv3}
\jname{}
\jinfo{}
\renewcommand{\footmark}{}

\usepackage{hyperref}
\usepackage[dvipsnames,usenames]{xcolor}
\definecolor{darkblue}{rgb}{0, 0, 0.5}
\hypersetup{colorlinks=true,citecolor=darkblue, linkcolor=darkblue, urlcolor=darkblue}

\usepackage{multirow}
\usepackage{booktabs}
\usepackage{amsmath}
\usepackage{framed}
\usepackage{enumitem}
\usepackage{pdfpages}
\usepackage{amsmath,soul}
\usepackage{soulpos}
\usepackage{setspace}
\ulposdef{\ulnumaux}{%
   $\underset{\saveulnum}{\rule[-.7ex]{\ulwidth}{.4pt}}$}

\newcommand{\ulnum}[2]{%
  \def\saveulnum{#1}%
  \ulnumaux{#2}}

\usepackage[utf8]{inputenc}
\newcommand{\DIRECT}[0]{{\small {\sf DIRECT}}}
\newcommand{\QUASI}[0]{{\small {\sf QUASI}}}
\newcommand{\NOMASK}[0]{{\small {\sf NO\_MASK}}}
\newcommand{\CODE}[0]{{\small {\sf CODE}}}
\newcommand{\DATETIME}[0]{{\small {\sf DATETIME}}}
\newcommand{\DEM}[0]{{\small {\sf DEM}}}
\newcommand{\LOC}[0]{{\small {\sf LOC}}}
\newcommand{\MISC}[0]{{\small {\sf MISC}}}
\newcommand{\ORG}[0]{{\small {\sf ORG}}}
\newcommand{\PERSON}[0]{{\small {\sf PERSON}}}
\newcommand{\QUANTITY}[0]{{\small {\sf QUANTITY}}}
\newcommand{\HEALTH}[0]{{\small {\sf HEALTH}}}
\newcommand{\POLITICS}[0]{{\small {\sf POLITICS}}}
\newcommand{\SEX}[0]{{\small {\sf SEX}}}
\newcommand{\BELIEF}[0]{{\small {\sf BELIEF}}}
\newcommand{\NOTCONF}[0]{{\small {\sf NOT\_CONFIDENTIAL}}}
\newcommand{\ETHNIC}[0]{{\small {\sf ETHNIC}}}
\newcommand{\CARDINAL}[0]{{\small {\sf CARDINAL}}}
\newcommand{\ENT}[1]{{\small {\sf #1}}}

\usepackage{stackengine}
\usepackage{dashrule}

\usepackage{dashrule}
\newlength\lunderset
\usepackage{setspace}

\issue{1}{1}{2016}

\runningtitle{The Text Anonymization Benchmark (TAB)}

\runningauthor{Ildikó Pilán et al.}

\begin{document}

\title{The Text Anonymization Benchmark (TAB):\\ A Dedicated Corpus and Evaluation Framework for Text Anonymization}

\author{Ildikó Pilán\thanks{The two first authors contributed equally to this work.}}
\affil{Norwegian Computing Center,\\ Oslo, Norway}

\author{Pierre Lison$^*$}
\affil{Norwegian Computing Center,\\ Oslo, Norway}

\author{Lilja Øvrelid}
\affil{Language Technology Group, University of Oslo, Norway}

\author{Anthi Papadopoulou}
\affil{Language Technology Group, University of Oslo, Norway}

\author{David Sánchez}
\affil{Universitat Rovira i Virgili, CYBERCAT, UNESCO Chair in Data Privacy, Spain}

\author{Montserrat Batet}
\affil{Universitat Rovira i Virgili, CYBERCAT, UNESCO Chair in Data Privacy, Spain}

\maketitle

\begin{abstract}
We present a novel benchmark and associated evaluation metrics for assessing the performance of text anonymization methods. Text anonymization, defined as the task of editing a text document to prevent the disclosure of personal information, currently suffers from a shortage of privacy-oriented annotated text resources, making it difficult to properly evaluate the level of privacy protection offered by various anonymization methods. This paper presents TAB (Text Anonymization Benchmark), a new, open-source annotated corpus developed to address this shortage. The corpus comprises 1,268 English-language court cases from the European Court of Human Rights (ECHR) enriched with comprehensive annotations about the personal information appearing in each document, including their semantic category, identifier type, confidential attributes, and co-reference relations. Compared to previous work, the TAB corpus is designed to go beyond traditional de-identification (which is limited to the detection of predefined semantic categories), and explicitly marks which text spans ought to be masked in order to conceal the identity of the person to be protected. \\
Along with presenting the corpus and its annotation layers, we also propose a set of evaluation metrics that are specifically tailored towards measuring the performance of text anonymization, both in terms of privacy protection and utility preservation. We illustrate the use of the benchmark and the proposed metrics by assessing the empirical performance of several baseline text anonymization models. The full corpus along with its privacy-oriented annotation guidelines, evaluation scripts and baseline models are available on: \\ \url{https://github.com/NorskRegnesentral/text-anonymization-benchmark}

\end{abstract}

\section{Introduction}

Privacy is a fundamental human right and plays a critical role in the proper functioning of democratic societies. It is, among others, a key factor enabling the practice of informed and reflective citizenship \cite{cohen2012privacy} and protects individuals against threats such as social control, mass surveillance, censorship, and loss of autonomy and human dignity \cite{kasper2007privacy,SANTANEN20195,finn2013seven}. As such, the right to privacy is enshrined in the Universal Declaration of Human Rights (Art.~12) and is further articulated in multiple national and international legal instruments. This right to privacy is, however, increasingly challenged by the steadily growing volumes of online data that may be collected on various individuals. This led to the introduction of several regulatory standards -- most notably the General Data Protection Regulation (GDPR) in place in Europe since 2018 \cite{GDPR} -- specifically focused on issues related to data privacy. 

Those regulations impose a number of constraints on the storage, processing and distribution of data including personal information. One of those constraints is that personal data cannot be distributed to third parties (or even used for other purposes than the one employed when collecting the data) without a proper legal ground, such as the explicit and informed consent of the individuals to whom the data refers. However, obtaining the informed consent of all individuals can be problematic, particularly when the dataset offers no practical means of contacting those individuals or when the sheer volume of the data makes it infeasible. This difficulty has far-reaching consequences on scientific fields such as medicine, psychology, law, and social sciences in general \cite{rumbold2017effect,peloquin2020disruptive}, as those fields all heavily depend on being able to access and scrutinize datasets including personal information, as for instance electronic health records, interview transcripts or court cases.

An alternative to obtaining the explicit consent of all individuals is to apply {\it anonymization techniques}
so that data can no longer be \emph{unequivocally} associated to the individuals they refer to. By its very definition and, as stated in the GPDR, the output of a (successful) anonymization process no longer qualifies as personal data, and can as such be freely shared to third parties such as research organizations. 
For traditional, structured datasets stored in tabular formats (as in relational databases), anonymization can be enforced through well-established privacy models such as $k$-anonymity and its extensions \cite{Samarati,Samarati_Sweeney:1998,Li:2007}, or differential privacy \citep{Dwork2006a,10.1561/0400000042}, which all provide formal privacy guarantees. Masking techniques commonly employed to enforce those privacy models include data suppression, generalization, noise addition or micro-aggregation \cite{hundepool2012statistical}. 

The anonymization of unstructured data such as text documents is, however, a much more challenging task for which many open questions remain \cite{batet2018semantic,acl2021}. Sequence labeling techniques are often employed to detect and remove specific categories of (named) entities that may enable re-identification, such as names, phone numbers or addresses. 
However, because potentially re-identifying information is unbounded, these tools miss many less conspicuous textual elements -- such as mentions of a person's physical appearance, current profession or political opinions -- which may contribute to increasing the risk of disclosing the identity of the person(s) in question. 
For this reason, the kind of protection these approach achieve go under the term of \textit{de-identification} \cite{Ferrandez,dernoncourt2017identification,10.1093/idpl/ipx020}, \emph{i.e.,} the removal of information that \emph{directly} identifies a subject, but they do not qualify as \textit{anonymization}. As 
defined by GDPR and similar regulations, anonymization requires removing or masking \emph{any} information that individually/directly or in aggregate/indirectly may re-identify the subject. 

As a consequence of this prevalence of de-identification methods in NLP, most existing evaluation benchmarks related to privacy protection focus on de-identification rather than anonymization. They may as a result overestimate the actual level of privacy protection achieved by the methods proposed in the literature.  

\subsection*{Contributions}

To remedy this situation, this paper presents TAB (Text Anonymization Benchmark), a corpus designed to evaluate the level of privacy protection offered by text anonymization methods. TAB consists of a collection of 1,268 court cases from the European Court of Human Rights (ECHR) in which each personal information expressed in the documents is explicitly annotated and associated with various properties, such as its semantic category, identifier type, confidential attributes and co-reference relations. Crucially, the annotation process was approached as an actual anonymization task. In particular, rather than merely annotating text spans of certain semantic types (e.g., names, locations, organizations), as done in virtually all previous works (see Section \ref{related}), the annotation was explicitly focused on identifying textual elements that may affect the disclosure risk of the individual to protect. This annotation process seeks to reflect the way in which human experts approach manual document sanitization in practice \cite{Bier}. Each text span annotated in the TAB corpus is associated with a \textit{masking decision} that expresses whether, according to the annotator, the span ought to be masked in order to conceal the identity of the person in question. To our knowledge, the TAB corpus constitutes the first publicly available text corpus for evaluating privacy-protection methods that goes beyond de-identification and specifically targets the (arguably harder) problem of text anonymization. 

In contrast with other privacy-oriented datasets and corpora, which are mostly framed in the medical domain \cite{meystre2010automatic,aberdeen2010mitre}, the TAB corpus is based on texts extracted from court cases, which are particularly appealing for the development and evaluation of general-purpose text anonymization methods. In particular, while medical records often exhibit a relatively narrow set of personal identifiers, court cases contain rich and unconstrained biographical descriptions of real subjects (plaintiffs, witnesses and other parties involved in the legal dispute), along with detailed depictions of events those subjects have been involved in. These documents therefore incorporate a wide range of linguistic expressions denoting direct and indirect identifiers, including names, spatio-temporal markers, demographic traits and other personal characteristics that may lead to re-identification.

Along with the annotated corpus, we also propose a set of new \textit{evaluation metrics} that assess the level of privacy protection and utility preservation achieved by anonymization methods more accurately than the standard IR metrics employed in the literature (see Section \ref{metrics}). In particular, the proposed privacy metrics operate at the level of entities rather than occurrences, capturing the fact that a personal identifier is only concealed to the reader if all of its occurrences in a given document are masked. Those metrics also account for the fact that personal identifiers are not all equally important -- in particular, disclosing a direct identifier such a full person name is a more serious privacy threat than more indirect information such as the person's nationality or gender. Finally, our utility metric considers the information lost as a result of masking by measuring the amount of information conveyed by masked terms.


The third and final contribution of this paper is the application of proposed benchmark and evaluation metrics to assess the level of privacy protection and utility preservation achieved by several anonymization methods. 
In particular, we provide baseline results for three types of approaches:
\begin{enumerate}
    \item A generic neural model trained for named entity recognition (NER),
    \item A privacy-oriented NER-based text de-identification system,
    \item Sequence labeling models based on large, pre-trained language models (BERT) fine-tuned on court cases from the TAB corpus. 
\end{enumerate}

The evaluation results obtained with these baseline approaches demonstrate the difficulty of the text anonymization task, and the limitations of traditional, NER-oriented de-identification methods in regard to preventing identity disclosure. 


\subsection*{Plan}

The remainder of the paper is organized as follows: 
\begin{itemize}
    \item Section \ref{sec:anonymization} provides a more detailed definition of the text anonymization problem and contrasts it with other privacy-enhancing techniques. 
    \item Section \ref{related} summarizes and discusses the limitations of current datasets employed to evaluate text anonymization methods. 
    \item Section \ref{corpus} presents our evaluation corpus and the annotation guidelines. 
    \item Section \ref{analysis} provides a quantitative analysis of the annotation process, including inter-annotator agreement. 
    \item Section \ref{metrics} presents the evaluation metrics we propose to measure the disclosure risks and preserved utility of anonymized texts. 
    \item Section \ref{results} reports and discusses evaluation results for three distinct text anonymization methods. 
    \item Finally, Section \ref{conclusion} gathers the conclusions and outlines future research directions.

\end{itemize}

\section{Background}
\label{sec:anonymization}

Privacy is often defined as the ability for individuals or groups to selectively withhold information about themselves~\citep{Westin}. Various regulations have been introduced to uphold this right to privacy in the digital sphere, and stipulate how personal data (that is, any information relating to an identified or identifiable person) may be collected and used. Although privacy regulations such as the General Data Protection Regulation (GDPR) in Europe, the California Consumer Privacy Act (CCPA) in the United States or China's Personal Information Protection Law (PIPL) have important differences in both scope and implementation, they all rest on the idea that data owners must have a valid \textit{legal ground} to be allowed to store, process, or share personal data\footnote{The most common legal ground is the explicit consent of the data subjects, but data owners can also invoke other grounds, such as the necessity to process data due to legal or contractual obligations.}. Datasets including personal data cannot be distributed to third parties without such a legal ground, as this would impair the privacy of the data subjects.

\subsection{The anonymization task}

Datasets can, however, be \textit{anonymized} to ensure they can no longer be attributed to specific individuals, in which case they fall outside the scope of privacy regulations. 
Anonymization is the \textit{complete} and \textit{irreversible} removal from a dataset of all information that, directly or indirectly, may lead to an individual being \emph{unequivocally} re-identified. Re-identifying information can therefore fall into one of the following categories: 

\begin{itemize}

    \item {\it Direct identifiers} correspond to values that are unique to a given individual, and can therefore directly disclose their identity. Examples of such direct identifiers include the full name of a person, their cellphone number, address of residence, email address, social security number, bank account, medical record number, and more.

\item {\it Quasi-identifiers} (also called \textit{indirect} identifiers) correspond to publicly known information on an individual (\emph{i.e.,} background knowledge) that does not enable re-identification when considered in isolation, but may do so when combined with other quasi-identifiers appearing in the same context.
For instance, the combination of gender, birth date and postal code can be exploited to unequivocally identify between 63 and 87\% of the U.S. population, due to this information being public availability int the US Census Data \citep{golle2006}. Quasi-identifiers encompass a broad range of semantic categories such as demographic characteristics of the individual, temporal or geographic markers, and their possible types are considered to be unbounded \cite{hundepool2012statistical}. Examples of quasi-identifiers are gender, nationality, name of employer, city of residence, date of birth (or other dates associated with the individual), personal acquaintances, number of criminal convictions, places the individual has visited in the past, and many more. 
\end{itemize}

Since removing direct identifiers is not sufficient to preclude re-identification, anonymization also necessitates to mask (i.e.~remove or generalize) quasi-identifiers. These operations necessarily entail some loss of information or data utility, as part of the document's content must be deleted or replaced by less specific text. As the ultimate objective of anonymization is to produce useful datasets that can be employed by third parties for purposes such as scientific research, the best anonymization methods are those that optimize the trade-off between minimizing the disclosure risk and preserving as much data utility as possible.
In fact, because ascertaining which information may serve as quasi-identifier (either manually or automatically) can be prone to errors or omissions, one would usually enforce anonymization beyond preventing strict unequivocal re-identification, and would aim at reducing the re-identification risk as much as the utility needs of the anonymized outcomes permit.

Unfortunately, very few legal frameworks have concretized the broad definition of anonymization as lists of (quasi-)identifiers.
The Health Insurance Portability and Accountability Act (HIPAA) in the United States~\citep{HIPAA} is a example of such legal framework, which defines 18 types of (quasi-)identifying information that can be typically found in medical documents. Protecting healthcare data according to such 18 types is considered legally compliant anonymization in the U.S., even though these types do not constitute, by any means, an exhaustive list of quasi-identifiers. As a result, HIPAA-based protection should be accounted for \emph{de-identification} rather than \emph{anonymization}. Other regulations such as the GDPR acknowledge this problem and define a tighter notion of anonymization, but this means that (quasi-)identifying information needs to be carefully assessed on dataset/document basis.
Consequently, the anonymization of text documents must consider how \textit{any} textual element may affect the disclosure risk, either directly or through semantic inferences, based on background knowledge assumed to be available to an adversary seeking to uncover the identity of the individuals referred to in the document \cite{batet2018semantic,acl2021}. 

As quasi-identifiers cannot be limited to a fixed set of semantic categories, de-identification approaches have been criticized for not masking \textit{enough} information to prevent re-identification \cite{batet2018semantic,acl2021}. Paradoxically, they may also remove \textit{too much} information. De-identification methods indeed systematically mask all occurrences of a given semantic type (such as date or location) without regard to their actual impact on the disclosure risk of the individual to be protected. As demonstrated in the empirical analysis of the TAB corpus (see Section \ref{analysis}), a substantial proportion of entities falling into the categories considered by de-identification methods may actually be left in clear text without noticeable impact on the disclosure risk. 


\subsection{Text anonymization techniques}

Approaches to text anonymization can be divided into two independent families. On the one hand, NLP approaches often rely on sequence labeling and formalize this task as a variant of Named Entity Recognition (NER) \cite{Chi:Nic:16,Lam:Bal:Sub:16}, where the entities to detect correspond to personal identifiers. Most existing work in this area has been focused on the medical domain, for which the existence of the HIPAA safe harbor rules
facilitates and standardizes the task. Indeed, PHI markers can be detected using rule-based and machine learning-based methods, either alone or in combination \cite{sweeney1996replacing,neamatullah2008automated,YANG2015S30,Yog:May:Pfa:2018}. Various neural architectures have also been proposed for this task and have been shown to achieve state-of-the-art performance, using e.g.~recurrent neural networks with character embeddings \cite{dernoncourt2017identification,liu2017identification} or bidirectional transformers \cite{johnson2020deidentification}. Section \ref{related} describes in more details the corpora and evaluation methodologies that are employed in clinical NLP to perform such de-identification.  

The second type of text anonymization methods relies on on privacy-preserving data publishing (PPDP). In contrast to NLP approaches, PPDP methods \cite{Ksafety,Kconfu,tPlaus,Sanchez2016,Sanchez2017} operate with an explicit account of {\it disclosure risk} and anonymize documents by enforcing a privacy model. 
As a result, PPDP approaches are able consider any term that may re-identify a certain entity to protect (a human subject or an organization), either individually for direct identifiers (such as the person's name or a passport) or in aggregate for quasi-identifiers (such as the combination of age, profession and postal code). This process will often depend on the background knowledge that is assumed to be available to an adversary -- for instance, the $C$-sanitize paradigm of \cite{Sanchez2016,Sanchez2017} operates on the assumption that this background knowledge is the set of all web pages that are publicly available and indexed through search engines. PPDP approaches then frame the text anonymization problem as a search for the minimal set of masking operations (such as data suppression or generalization) on the document to ensure the requirements derived from the privacy model are fulfilled. Although these methods offer more formal and robust privacy guarantees than those based on sequence labeling, they also have a number of limitations and scalability issues \cite{acl2021}. In particular, PPDP approaches typically reduce documents to collections of terms and thereby ignore how terms are influenced by their context of occurrence. 

A common challenge faced by researchers working on text anonymization is the lack of a standard benchmark to evaluate and compare those anonymization methods. A widespread approach is to rely on human annotators to manually mark predefined types of personal information in a collection of documents, and then compare the system output with human annotations using IR-based metrics such as precision, recall and $F_1$ score. 

\subsection{Relation to other privacy-enhancing tasks}
\label{sec:other_privacy_tasks}

Privacy regulations such as GDPR, CCPA, PIPL or even HIPAA primarily focus on preventing {\it identity disclosure}, which occurs when a record (or, in the case of text anonymization, a document) in the anonymized dataset can be linked to a specific individual. However, personal confidential information may also be disclosed without re-identification. This phenomenon, called {\it attribute disclosure}, occurs when the released data can be exploited to unequivocally infer the value of a confidential attribute (e.g.~a criminal conviction) for a group of anonymized individuals with some shared characteristics. For instance, if all court cases related to individuals of a particular nationality end up with the same court verdict, we can infer the verdict of any person of that nationality (provided we know the presence of that person in the dataset) even though we are unable to link each individual with a specific court case.

This problem of attribute disclosure has been investigated in several NLP studies, in particular for the purpose of obfuscating documents to conceal sensitive social categories such as gender \cite{reddy2016obfuscating} or race \cite{blodgett2016demographic}. Recent deep learning approaches have sought to transform latent text representations (word or document embeddings) to protect confidential attributes using adversarial learning \cite{elazar-goldberg-2018-adversarial,barrett-etal-2019-adversarial}, reinforcement learning \cite{mosallanezhad-etal-2019-deep} or encryption \cite{huang-etal-2020-texthide}. Those methods, however, operate at the level of latent vector representations and do not modify the texts themselves. One notable exception is the text rewriting approach of \cite{xu2019privacy} which edits the texts using back-translations combined with adversarial training and approximate fairness risk. 

Several authors~\cite{li-etal-2018-towards,DBLP:conf/post/FernandesDM19,feyisetan2019leveraging} have also proposed privacy-preserving methods that focus on obfuscating the author(s) of the document rather than protecting the privacy of the individuals referred to in the text. The authorship of a document and the author's attributes are inferred from the linguistic and stylistic properties of the text rather than the document's topic or the text semantics. Those approaches rely on distorting the distribution of words by inserting differentially private noise to the word embeddings~\cite{DBLP:conf/post/FernandesDM19,feyisetan2019leveraging} or constraining the embeddings to prevent disclosing certain attributes~\cite{li-etal-2018-towards}. The outputs of those systems are therefore typically distorted bag-of-words or distributed word representations rather than actual documents. 

Differential privacy \citep{Dwork2006a,10.1561/0400000042} has also been employed for other privacy-oriented NLP tasks such as producing synthetic texts \cite{neurips2019_synthesis}, producing text transformations able to resist membership inference attacks\footnote{Membership Inference Attacks aim to identify whether a data sample (such as a text) was part of the training set employed for learning a machine learning model \cite{shokri2017membership}.}  \cite{krishna-etal-2021-adept,Habernal.2021.EMNLP} or learning deep learning models with privacy guarantees \cite{mcmahan_learning_2017,li2021large}. 

Those approaches all provide valuable privacy-enhancing techniques that make it possible to create texts (or text representations) that are oblivious to certain demographic attributes or hide the identity of the author. However, they seek to address a different task than text anonymization, which focuses on preventing identity disclosure by masking the personal identifiers expressed in the text. In particular, the objective of text anonymization is to produce a modified version of a document where re-identifying information is masked through suppressions and generalizations of text spans, but without altering the parts of the document that do not correspond to (direct or indirect) personal identifiers.

This need to preserve the semantic content conveyed in the text is an important prerequisite for most types of data releases. For instance, medical records in which the clinical observations have been randomly altered are of little use to medical researchers, who need a guarantee that the anonymization preserves the ``truthfulness'' of the initial record, in this case the description of medical symptoms and resulting diagnosis. For instance, a medical condition such as ``bronchitis'' may be replaced/masked in a medical record by a generalization such as ``respiratory disease'', but not by any other disease (being respiratory or not), because the latter may mislead medical researchers. The same argument applies for court cases where it is desirable to protect the identity of certain individuals (plaintiffs, witnesses, victims), but where the actual judgment, even though it could be made less specific, should not be modified, lest the resulting text becomes useless for legal professionals. In other words, the masking process for those types of data releases should be non-perturbative: each masked term should strictly encompass a subset of the semantics of the original term, which implies that each masked term must be either concealed or replaced by generalizations. Although there exists a number of differentially-private text transformation methods such as ADePT \cite{krishna-etal-2021-adept} that seek to minimize the amount of perturbations introduced on certain properties of the text, such as the ability to determine the general intent of an utterance, they do effectively produce new texts (or text representations) rather than truthful, masked versions of existing documents. 


\section{Related benchmarks}
\label{related}

\subsection{Medical datasets}
Most existing NLP studies on text anonymization have been performed in the area of clinical NLP, where the goal is to detect PHI entities in medical texts \cite{meystre2010automatic,aberdeen2010mitre}. 
Several shared tasks have contributed to increased activity within this research area, in particular through the release of evaluation datasets for text anonymization manually annotated with PHIs. 
Most notable are the 2014 i2b2/UTHealth shared task \cite{stubbs2015annotating} and the 2016 CEGS N-GRID shared task \cite{stubbs2017identification}.

The 2014 i2b2/UTHealth shared task \cite{stubbs2015annotating} is composed of patient medical records annotated for an extended set of PHI categories. The training/evaluation dataset contains 1,304 longitudinal medical records describing a total of 296 patients. The authors applied what they termed a "risk-averse interpretation of the HIPAA guidelines" which expanded on the set of categories to include indirect identifiers that could be used in combination to identify patients. These include names of hospitals, doctors and nurses, patient’s professions, as well as various expressions related to dates, age, and locations. They further adopted a hierarchical annotation scheme with fine-grained sub-categories indicating e.g., identification numbers (social security number, medical record number, etc) or parts of a location (room number, hospital department, street name). The annotation effort also focused on the generation of surrogate terms, a process which was largely guided by the fine-grained categorization. The annotation was performed by two annotators in parallel followed by an adjudication phase. Inter-annotator agreement was measured both at the entity and token level and the authors report a (micro) $F_1$ of 0.89 and 0.93 for entity and token level agreement, respectively. Current state-of-the-art performance on this dataset is achieved with fine-tuned pre-trained transformer language models such as BERT \cite{devlin2018bert} and its domain-specific variants, SciBERT \cite{beltagy2019scibert} and BioBERT \cite{lee2020biobert}, to achieve over 0.98 $F_1$ score \cite{johnson2020deidentification}.

In a follow-up to the 2014 task, the 2016 CEGS N-GRID shared task \cite{stubbs2017identification} released a training and evaluation dataset based on psychiatric intake records, which are particularly challenging to de-identify due to a higher density of PHIs. The aim of this shared task was to evaluate the extent to which existing de-identification systems generalize to new clinical domains. The annotation was performed over 1,000 intake notes using the annotation guidelines developed for the 2014 i2b2/UTHealth shared task described above. For this dataset, the inter-annotator agreement was measured at an entity level $F_1$ of 0.85 and token level $F_1$ of 0.91.



Some annotation efforts are also geared towards de-identification for languages other than English. For Swedish, \namecite{VELUPILLAI2009e19,Alf:Bri:Dal:2012} present efforts to collect and standardize annotated clinical notes. For Spanish, a recently held shared task on clinical de-identification released a synthetic Spanish-language dataset \cite{Mar:Gon:Int:2019}.

\subsection{Non-medical datasets}

Evaluation datasets for text anonymization outside the medical domain are usually small, shallowly annotated (focusing on de-identification rather than anonymization), already (pseudo)anonymized and/or not public. Unsurprisingly, the main reason for the latter is because of privacy concerns: an ideal evaluation dataset for text anonymization should relate to real-world individuals, and would therefore contain identifying and confidential personal features.

A common source for evaluation data are personal emails. The Enron email dataset\footnote{\url{https://www.cs.cmu.edu/~enron/}} is probably the most well-known, consisting of 0.5 million messages from 150 employees of Enron. A similar dataset, annotated w.r.t. a set of predefined identifying information types (names, addresses, organizations, etc.), was presented in \namecite{medlock2006introduction}. \namecite{Ede:Kri:Hah:2020} also present a dataset consisting of NER-oriented annotated German e-mails. Even though these datasets are publicly available, they bear an important limitation: due to the intrinsically sensitive nature of the original data, the released data have been already subjected to (pseudo)anonymization. This makes the evaluation carried out on these data less realistic, because the most disclosive information has been already redacted or pseudoanonymized. 

Other privacy-oriented corpora that are also limited in some way include a large corpus of SMS messages, already subjected to anonymization based on predefined dictionaries of sensitive terms \cite{patel2013}, a non-released NER-oriented Portuguese-English dataset of legal documents \cite{bick2015} or recently released pseudonymized Swedish learner corpus with restricted access \cite{megyesi2018learner}.

More recently, \namecite{Jensen2021DeidentificationOP} presented an annotated dataset based on job postings from StackOverflow. The dataset is large and openly available, but the text is semi-structured (rather than being plain text) and the annotation is limited to the NER-oriented categories: organization, location, contact, name, and profession.

As an alternative to using private or personal data, several approaches within the PPDP paradigm have employed Wikipedia biographies for evaluation purposes \cite{Chow,staddon2007,Sanchez2016,Sanchez2017,acl2021,tkde,wikipii}. Key motivations for this choice are the public availability of the texts (therefore not subjected to privacy issues), and the high density and large variety of (quasi-)identifying information they contain. Compared to the approaches discussed so far, which focus on NER-oriented annotations and de-identification rather than anonymization, these works operate on all terms that may cause direct or indirect re-identification of the individual to be protected, regardless of their semantic type. These privacy-oriented annotations are more accurate than NER-based annotations \cite{acl2021}, and better capture the way in which manual document redaction is done by human experts \cite{Bier}. However, annotations of text spans are just binary (either sensitive or not), and neither describe the entity type nor the type of incurred risk (identity or attribute disclosure). Moreover, those approaches only annotate a small collection of biographies (from 5 to 50), many of them without clear annotation guidelines or public release of the annotations -- although see \cite{acl2021,tkde} for two exceptions. 

Even though Wikipedia biographies are undeniably useful to evaluate the effectiveness of domain-independent anonymization methods, they are also highly copied texts: just searching for some exact extracts of text from Wikipedia articles in a web search engine returns many sources that have exactly reproduced such text. This may give the impression that some (very specific) information is more common that what it really is, because it is included and referred in many (copied) sources. This hampers the assessments made by distributional and information-theoretic anonymization methods \cite{Chow,Relationships,sanchez2013,Sanchez2016}, which specifically rely on web-scale statistics to assess disclosure risks. Furthermore, the fact that Wikipedia biographies are limited to public or ``notable'' personalities also introduces a substantial bias in the evaluation process, as it is often much easier to extract biographical details about those personalities than for average, non-public individuals. 

Large datasets containing personal and privacy-sensitive documents have also been created to evaluate data loss prevention (DLP) methods \cite{Vartanian,Hart,Trieu}. Even though DLP methods do assess the sensitivity of the information contained in textual documents, they only do it at the document level. Their goal is to design metrics to assess the sensitivity of a document and, from this, derive policies that can prevent or mitigate the effects of possible data leakages. Consistently, evaluation datasets for DLP just tag documents as sensitive or non-sensitive (or, at most, into several degrees of sensitivity); therefore, these annotations are not useful to properly evaluate anonymization methods.
Moreover, using these documents as source to create datasets for anonymization methods (by manually annotating their contents) may be ethically questionable, as many of those documents originate from data leakages such as Wikileaks\footnote{\url{https://www.wikileaks.org/}} or confidential archives leaked by the whistleblower Edward Snowden \footnote{\url{https://github.com/iamcryptoki/snowden-archive}}.

\section{The Text Anonymization Benchmark (TAB)}
\label{corpus}

The previous section highlighted two important limitations of current datasets for privacy evaluation of text data, namely that most of them are (1) restricted to clinical texts and (2) largely focus on NER-based de-identification rather than anonymization. Moreover, a common factor of previous annotation efforts is that each and every document in the dataset exclusively describes a single individual, such as a electronic health record associated to a specific patient. This setup considerably simplifies the anonymization process (both for manual annotations and when using automated methods), as one can assume that all the entities and facts mentioned in the document are directly related to the individual to protect. However, it also makes the task less realistic, as many types of text documents do refer to multiple individuals. 

Considering the limitations of the above described datasets, we searched for a document collection satisfying the following criteria:
\begin{itemize}
    \item It should contain rich, detailed documents in plain text format, rather than semi-structured data or short texts.
    \item It should contain generic personal information about real-world individuals. This personal information should encompass a large variety of direct and quasi-identifiers (not limited to predefined categories), including biographical details, demographic traits and depiction of events featuring temporal and geographic markers. 
    \item It should be based on public text sources that can be freely re-distributed, and should not have been subjected to any prior (pseudo)anonymization. It should also relate to a broad spectrum of individuals, and not only public figures, as as the case for e.g. Wikipedia.
\end{itemize}

The TAB corpus presented in this paper satisfies the above criteria. The corpus is based on court cases from the European Court of Human Rights (ECHR), which is an international court of the Council of Europe\footnote{See \url{https://www.echr.coe.int/}}. Its purpose is to interpret the European Convention on Human Rights, an international convention adopted by the Council of Europe in 1950 and designed to protect human rights and political freedoms across Europe \cite{gearty1993european}. The court rules on applications relating to breaches of the rights enumerated in the Convention. As of 2020, the court disposes judicially of about 40,000 applications every year, covering cases originating from any of the 47 countries that are part of the Council of Europe and have ratified the Convention. The court's working languages are English and French, and all court cases are publicly available in full-text on the court website. Their publication has received the consent of the applicants\footnote{By default, documents in proceedings before the Court are public, and applicants wishing to bring a case before the Court are informed of this requirement. It is, however, possible to submit a request for anonymity, either along with the application or retroactively, cf. {\it Rules of Court}, Rules 33 and 47.}.

\subsection{Preprocessing}

The selection of court cases to include in the TAB corpus followed a number of criteria. First, we only included English-language judgments in the corpus, leaving aside the French-language judgments. We ignored judgments that had been anonymized prior to publication (which happens for especially sensitive cases, among others when the case involves children), and also filtered out judgments released after 2018 (as applicants have the possibility to submit to the Court retroactive requests to anonymize the court proceedings). Furthermore, we only included judgments from the ``Grand Chamber'' and ''Chamber'', leaving aside smaller judgments decided in Committees or Commissions, that often contain fewer and less interesting personal identifiers.

ECHR court cases are typically divided in 5 parts: \begin{enumerate}
    \item An introduction stating the alleged breach (by a contracting state) of at least one of the rights enumerated in the Convention.
    \item A ``Statement of Facts' structured as a list of factual elements (without legal arguments) that underpin the application.
    \item A ``Legal Framework`` enumerating national and international legal material (laws, previous judgments, jurisprudence) relevant to the case.
    \item A section entitled ``The Law'' which details the legal arguments put forwards by each party (applicants vs. government representatives), along with the reasoning followed by the Court.
    \item A conclusion stating the final judgment of the Court. 
\end{enumerate}

Most personal identifiers are typically found in the two first sections (Introduction and Statement of Facts). As a consequence, the annotation was restricted to those two sections, leaving aside the parts related to the legal interpretation. 




\subsection{Annotation process}
\label{sec:annotation}

The corpus was annotated by twelve university students at the Faculty of Law of the University of Oslo over a two-month period. The annotation was conducted using a web interface\footnote{See \url{https://www.tagtog.net/}}, and students received financial remuneration for their work. The annotators were given a detailed set of annotation guidelines (see Appendix). All annotators completed an initial training phase where they were instructed to annotate the same court cases, compare their outputs, and resolve any potential disagreements on the basis of the guidelines. 



The annotators are provided with the Introduction and Statement of Facts extracted from an ECHR court case, together with the name of a specific individual that should be protected in this document (see Fig. \ref{fig:annotation} for an example). The annotation process is thus focused on concealing the identity of one single person. Personal information pertaining to other individuals mentioned in the case should only be masked insofar they provide indirect cues that may enable the re-identification of the main person to protect.

The annotators are instructed to first read through the entire document and go through the following steps:
\begin{itemize}
    \item \textbf{Step 1} focuses on determining all phrases that contain personal information and classify them according to their semantic type.
    \item \textbf{Step 2} then looks at each entity mention marked in Step 1 and determines for each whether it can be kept in clear text or needs to be masked to conceal the identity of the person to protect -- and in this latter case, whether it corresponds to a direct identifier or a quasi-identifier.
    \item \textbf{Step 3} enriches the entity mentions with a second attribute indicating whether they correspond to confidential information (such as religious beliefs, ethnicity or health data).
    \item \textbf{Step 4} connects entity mentions that relate to the same underlying entity but do not have the same string value (such as a person name that may be written with or without the first name).
    \item Finally, in \textbf{Step 5}, the annotations undergo a last process of quality control where pairs of annotators review each other's documents to ensure that the identity of the person to protect is properly concealed. 
\end{itemize}

We detail each of these five steps below, and then describe the subsequent quality reviews that are applied to the annotation outputs (Section \ref{lbl:quality_check}). To ensure that the annotators had a good understanding of each court case and the background knowledge surrounding it, the annotators were assigned court cases for which the national language of the country accused of human rights violations was familiar to the annotator. For instance, court cases filed against Germany were annotated by law students with a working knowledge of German, such that they could more easily understand the general context behind the case. 

In total, the students used a total of around 800 hours on this annotation work (excluding the hours devoted to training and status meetings), leading to an average of 22 minutes to annotate and quality-check a single court case. 

\subsection*{Step 1: Entity Detection}
\label{semantic types}

In this step of the annotation process, the annotators are instructed to mark all text spans denoting some type of personal information and assign them to a semantic class. While this task is superficially similar to classical named entity annotation, the inventory of categories differs somewhat from commonly used NER schemes and is not restricted to proper nouns. In particular, entities providing demographic information will often be common nouns or even adjectives.

The entity types are the following:
\begin{description}
\item[\PERSON{}] Names of people, including nicknames/aliases, usernames and initials.
\item[\CODE{}] Numbers and identification codes, such as social security numbers, phone numbers, passport numbers or license plates
\item[\LOC{}] Places and locations, such as 
cities, areas, countries, addresses, named infrastructures etc.
\item[\ORG{}] Names of organizations, such as 
public and private companies, schools, universities, public institutions, prisons, healthcare institutions, non-governmental organizations, churches, etc
\item[\DEM{}] Demographic attributes of a person, such as native language, descent, heritage, ethnicity, job titles, ranks, education,physical descriptions, diagnosis, birthmarks, ages
\item[\DATETIME{}] Description of a specific date (e.g. {\it October 3, 2018}), time (e.g. {\it 9:48 AM}) or duration (e.g. {\it 18 years}).
\item[\QUANTITY{}] Description of a meaningful quantity, e.g. percentages or monetary values.
\item[\MISC{}] Every other type of personal information associated (directly or indirectly) to an individual and that does not belong to the categories above.
\end{description}

In this stage the annotators were instructed to mark all entities according to their type, without taking into account whether those entities need to be masked to protect the individual in question\footnote{The only exceptions from this rule were entities indicating the profession or title of the legal professionals involved in the case (for instance {\it solicitor}, {\it legal adviser}, {\it lawyer}, etc.) and parts of generic legal references (such as the year a particular law was passed or published).}. Country names were labeled \LOC{} when referring to the geographical location, but \ORG{} when referring to the government of that country or its representatives. 

To make the annotation process as effective as possible, the documents were provided to the annotators with a pre-annotation produced by combining an off-the-shelf named entity recognition tool (spaCy) with a set of handcrafted heuristics tailored for the recognition of common entities such as dates, codes and quantities. Annotators were instructed to carefully inspect the pre-annotations and validate, revise or remove them in accordance with the annotation guidelines. Statistical analysis of the resulting annotations showed that the annotators did substantial edits on the pre-annotations, with around 24 \% of all entity mentions that were either corrected from the initial pre-annotations or new entities added manually by the annotator. 

\subsection*{Step 2: Masking}

In the second phase of the annotation, the annotators were instructed to determine whether the entity mention ought to be masked to protect the individual in question. If the entity mention is to be masked, we further distinguish between {\it direct identifiers}, which can unequivocally lead to re-identification, and {\it quasi-identifiers}, which can lead to the re-identification of the individual to protect when combined with other quasi-identifiers along with background knowledge. 

For re-identification to be possible, quasi-identifiers must refer to some personal information that can be seen as potential “publicly available knowledge” — i.e. something that we can expect that an external person may already know about the individual or may be able to infer through legal means —, and the combination of quasi-identifying information should be enough to re-identify the individual with no or low ambiguity. The annotators were explicitly instructed to assess whether it is likely that a motivated adversary could, based on public knowledge, gain access to the quasi-identifying values of the individual to be protected. 

As a rule of thumb, immutable personal attributes (e.g., date of birth, gender or ethnicity) of an individual that can be known by external entities should be considered quasi-identifiers. Circumstantial attributes (such as the date or location of a given event mentioned in the court case) may be considered quasi-identifiers or not according to the chance that external entities may obtain knowledge about such information. For instance, the usual residence of a person or the date of a hospital admission can be expected to be known by third parties, while the exact time a person went to the grocery store will typically not. The annotators were also instructed to consider as public knowledge any information that can typically be found on the web.\footnote{There is, however, one important exception to this rule of viewing all web content as ``public knowledge''. The annotators were indeed instructed to regard the actual text of the court case as not part of public knowledge, although it is in practice available on the ECHR website and in some online legal databases. Without this exception, the anonymization process would become meaningless, as one can easily link back the anonymized text with its original version on the ECHR website (for instance by searching for the presence of a few phrases occurring in the text) and thereby re-identify the person, as demonstrated in \cite{weitzenboeck2022gdpr}.}

\subsection*{Step 3: Confidential attributes}
Annotators were in addition instructed to indicate whether an entity describes a confidential attribute, i.e. conveying information that, if disclosed, could harm or could be a source of discrimination for the individual to protect. Due to their confidential nature, those attributes are typically not known by external entities, and are therefore rarely seen as quasi-identifiers. They are, however, important to consider if one wishes to prevent attribute disclosure (see Section \ref{sec:other_privacy_tasks}).

The categories of confidential attributes follow the ``special categories of personal data'' defined in the GDPR\footnote{\url{https://ec.europa.eu/info/law/law-topic/data-protection/reform/rules-business-and-organisations/legal-grounds-processing-data/sensitive-data/what-personal-data-considered-sensitive_en}}:
\begin{description}
\item[\BELIEF{}] Religious or philosophical beliefs
\item[\POLITICS{}] Political opinions, trade union membership
\item[\SEX{}] Sexual orientation or sex life
\item[\ETHNIC{}] Racial or ethnic origin
\item[\HEALTH{}] Health, genetic and biometric data. This includes sensitive health-related habits, such as substance abuse
\item[\NOTCONF{}] Not confidential information (most entities)
\end{description}

\begin{figure}
    \centering
    \includegraphics[width=1.0\textwidth]{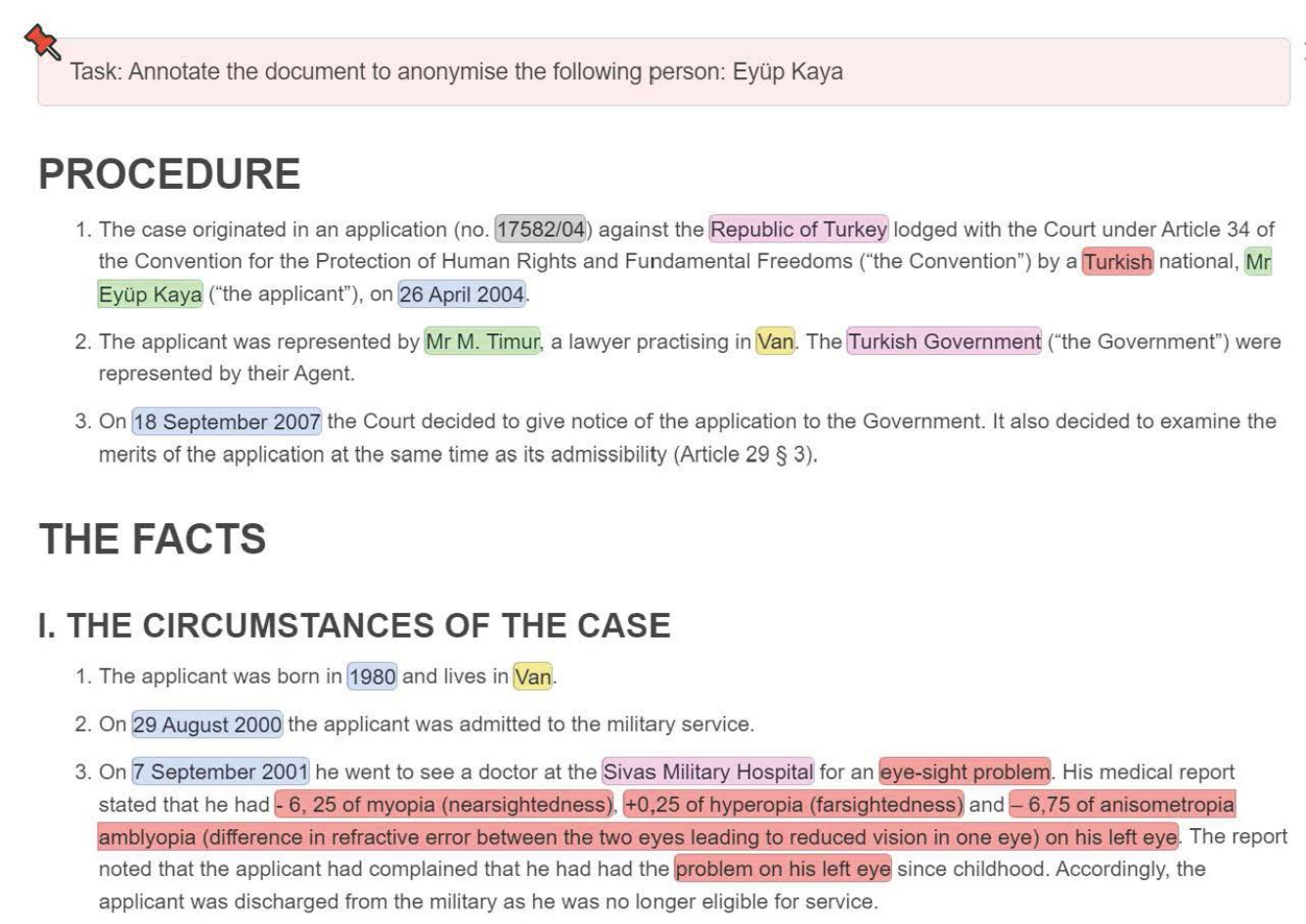} \\ $\phantom{c}$
    \caption{Example of court case displayed on the annotation interface (TagTog). The person to protect is specified in the task box on the top of the page. The text to annotate consists of the Introduction and Statement of Facts of the court case. Entities annotated during Step 1 are color-highlighted. Steps 2 and 3 are then carried out via a pop-up window (not shown on the Figure). }
    \label{fig:annotation}
\end{figure}

Figure \ref{fig:annotation} illustrates an example of ECHR court case displayed on the annotation interface. 

\subsection*{Step 4: Entity linking}
\label{sec:relations}

The protection of a nominal entity can only succeed if all mentions of this entity within the document are duly masked. However, the surface form of those mentions may vary, as for e.g. {\it John Smith} and {\it Mr Smith}, or {\it California Institute of Technology} and {\it CalTech}. To this end, we provide explicit co-reference relations between mentions referring to the same underlying entity. Annotators were instructed to explicitly mark relations between all mentions of the same entity within a given document. Entities with identical string values (e.g. {\it John Smith} and {\it John Smith}) were by default assumed to refer to the same entity, but this default choice could be overridden by the annotator. 

It should be noted that the relations between entity mentions only encompass a relatively small subset of the relations typically considered in co-reference resolution \cite{lee-etal-2017-end,SUKTHANKER2020139}. In particular, anaphoric expressions such as pronouns and possessive adjectives are not part of this annotation process, as they reveal little information about the individual to protect (with the possible exception of gender) and do not typically need to be masked. For instance, although {\it John Smith} and {\it he} may both refer to the same person, only the first expression is likely to increase the re-identification risk, while the second expression only indicates that the person is male. The entity linking step is therefore in practice limited to nominal and adjectival phrases that convey roughly the same information about their underlying entities and need to be considered as part of the masking process. 

\subsection*{Step 5: Quality reviews}
\label{lbl:quality_check}

\begin{figure}
    \centering
    $\ \ \ \ \ \ $\includegraphics[width=0.97\textwidth]{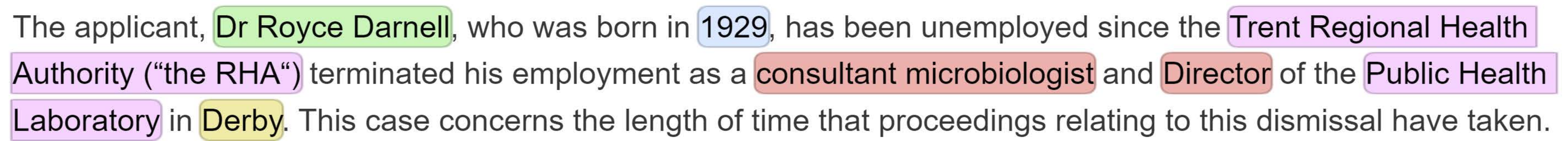} \\ $\phantom{c}$ \\[0mm]
    \includegraphics[width=0.9\textwidth]{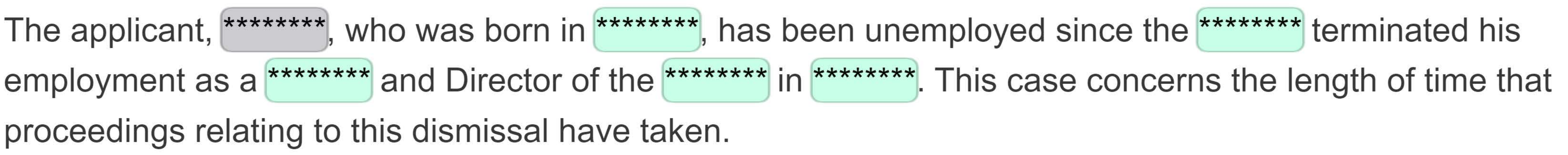}$\ \ \ \ \ \ $ \\ $\phantom{c}$
    \caption{Documents before (top) and after (bottom) entity masking based on annotator decisions. In this particular case, one entity was marked as direct identifier (``Dr Royce Darnell``), five entities as quasi-identifiers, and one entity (``Director'') was left in clear text.}
    \label{fig:masking}
\end{figure}

The final annotation phase consisted of a round of quality reviews for a subset of the annotated documents. Students were paired up for this purpose and were instructed to carefully review each other's annotations and assess whether the identity of the person specified in the anonymization task was sufficiently protected. 
To facilitate this review process, we generated a masked version for each annotated document, where all entities marked as direct or quasi-identifiers requiring masking were replaced by '*' symbols as shown in Figure \ref{fig:masking}.  

In case of doubt, students were instructed to validate the masking decisions by checking whether they could re-identify the person mentioned in the court case based on information available on the web. 

\subsection{Corpus release}

We release the corpus in a standoff JSON format. Each annotated text span is associated with the annotator name, its offset in the text, a semantic category, the identifier type (direct, quasi, or no need to mask), possible confidential attributes, and an unique identifier for the entity it refers to (based on the relations from Step 4).\footnote{ The data, its documentation and the annotation guidelines are available at \url{http://cleanup.nr.no/tab.html}.}

Even though the whole corpus can be used for evaluation purposes, we defined a training set (80\% of unique court cases), a development (10\%) and a test set (10\%), so that the corpus may also be used to build supervised anonymization models. As we have a varying number of annotators for each case, the corpus was divided as to maximize the proportion of court cases with multiple annotators in the development and test sets. Those two sets also only include court cases where at least one annotation was quality-reviewed as described above. Table \ref{tbl:split} provides some general statistics over the corpus split. 

\begin{table}[h]
\caption{\label{tbl:split} Statistics over the data splits.}
\begin{tabular}{ccccc}
    \toprule
        Split &
        \# docs &
        \# doc annotations & 
        \# reviewed annotations &
        avg. \# annotators \\
    \midrule
	    train   & 1,014 & 1,112 &	328 & 1.10 \\ 
        dev     & 127  & 541 & 313 & 4.26 \\ 
        test    & 127 &	555 & 315 & 4.37 \\ 
\end{tabular}
\end{table}

Text anonymization is a task that is not restricted to a single solution, at least when the text contains quasi-identifiers \cite{acl2021}. Indeed, it is precisely the combination of quasi-identifiers that may create a privacy hazard, and there may be several alternative sets of masking operations that may be applied on those quasi-identifiers to reduce the disclosure risk below an acceptable threshold. In contrast to the bulk of NLP resources where multiple annotations are often adjudicated manually or automatically \cite{boisen-etal-2000-annotating}, the TAB corpus therefore retains the masking choices of each annotator separately, without merging them into a single layer. 

The appendix contains a full example of a document annotated with entities and masking decisions.

\section{Data analysis}
\label{analysis}

\subsection{General statistics}

Table \ref{tbl:data} presents a quantitative overview of the annotated corpus. As Table \ref{tbl:data} shows, about 22\% of cases have been annotated by more than one annotator, with 163 cases having three or more annotators.  
The portion of quality reviewed document annotations according to the procedure described in Section \ref{lbl:quality_check} was 43\%.
Documents were 1442 tokens long on average. The table also distinguishes between the number of distinct entities and the number of mentions, as an entity (for instance a person or an organization) may be mentioned several times through a given document. 

\begin{table}[h]
\caption{\label{tbl:data} General statistics.}
\begin{tabular}{p{9cm}r}
    \toprule
    Number of distinct court cases (documents): & 1,268 \\
    Number of distinct document annotations: & 2,208 \\
    Number of documents with multiple annotators: & 274 \\
    Number of document annotations reviewed for quality by another annotator: & 956 \\
    Number of (distinct) annotated entities: & 108,151\\
    Number of annotated entity mentions: & 155,006 \\
    Total number of tokens: & 1,828,970 \\
    \toprule
\end{tabular}
\end{table}

\begin{table}[h] 

\caption{\label{tbl:entity_distr} Distribution of entity types across the TAB corpus, along with their corresponding identifier type (direct identifier, quasi identifier, or no need to mask) and confidential status. The parenthesis in the first column refers to the proportion of entities of that type, in percent. The parentheses in the three other columns refer to the percentage of entities within this entity type that are respectively labeled as direct identifier, quasi-identifier, or have a confidential status.}
\vspace{2mm}
\begin{tabular}{l|r|rr|r}
    \toprule
       Entity type & \# mentions & \# direct & \# quasi & \# confidential\\
    \midrule
        \DATETIME{} & 53,668 \ \ (34.6) & 23 \ (0.04) & 48,086 \ (89.6) & 530 \ (0.99)  \\
        \ORG{} & 40,695 \ \ (26.3) & 20 \ (0.05) & 12,880 \ (31.6) & 866 \ \ \  (2.1)  \\
        \PERSON{} & 24,322 \ \ (15.7) & 4,182 \ (17.2) & 15,839 \ (65.1) & 413 \ \ \  (1.7)  \\
        \LOC{} & 9,982 \ \ \ \ (6.4) & 1 \ (0.01) & 6,908 \ (69.2) & 19 \ \ \ (0.2)  \\
        \DEM{} & 8,683 \  \ \ \ (5.6) & 1 \ (0.01) & 4,166 \ (48.0) & 2,278 \ (26.2)  \\
        \MISC{} & 7,044 \  \ \ \ (4.5) & 28 \ \ (0.4) & 3,437 \ (48.8) & 1,125 \ (16.0)  \\
        \CODE{} & 6,471 \  \ \ \ (4.2) & 2,484 \ (38.4) & 3,558 \  (55.0) & 18 \ \ \ (0.3)  \\
        \QUANTITY{} & 4,141 \  \ \ \ (2.7) & 0  \ \ (0.0) & 3,370 \  (81.4) & 87 \ \ \ (2.1)  \\  \midrule
        \bf Total & 155,006 (100.0) & 6,739 \ \ (4.4) & 98,244 \  (63.4) & 5,336 \ \ \ (3.4)  \\
\end{tabular}
\end{table}

\begin{table}[t]
\caption{\label{tbl:conf_distr} Distribution of confidential attribute types.}
\begin{tabular}{lrr}
    \toprule
        \ Confidential status &
        \ Count & 
        \ \% \\
        \midrule
            \HEALTH{}           &       2,320  & 1.5 \\
            \POLITICS{}         &       1,039  & 0.7 \\
            \ETHNIC{}           &        806  & 0.5 \\
            \BELIEF{}           &        655  & 0.4 \\
            \SEX{}              &        516  & 0.3 \\ 
            \NOTCONF{} &     {\it 149,670}  & {\it 96.6} \\
\end{tabular}
\end{table}

Table \ref{tbl:entity_distr} reports the distribution of entity types as well as the proportion of direct, (masked) quasi-identifiers and confidential entities per entity category. Out of the 155,006 entity mentions, most belonged to the \DATETIME{}, \ORG{} and \PERSON{} categories. Only 4.4 \% of the annotated entities were marked as direct identifiers, while 63 \% were marked as quasi-identifiers, and 32 \% were left without any mask.

The entities that were most frequently masked (either as a direct or quasi-identifier) were \CODE{} (93\%), \DATETIME{} (90\%), \PERSON{} (82\%) and \QUANTITY{} (81\%). In contrast, less than half of \DEM{} entities were masked 
Annotators masked on average 67.9\% of entities with a standard deviation of 8.3 \% across different annotators, which indicates a certain degree of subjectivity in text anonymization. 
Only 3\% of entities belonged to one of the confidential attribute categories as Table \ref{tbl:conf_distr} shows. These were mostly \HEALTH{}, such as \textit{speech impediment, anorexia}; and \POLITICS{}, for example, \textit{communist, Liberal Party} with all other confidential categories accounting for less than 1000 mentions.

\subsection{Inter-annotator agreement}
\label{sec:iaa}
Table \ref{tbl:iaa} details the level of agreement observed for several types of annotation. The first column denotes the average observed agreement (AOA), that is, the proportion of annotated items raters agreed on. We also include two other inter-annotator agreement (IAA) measures that correct for chance agreement: Cohen's $\kappa$ and Krippendorff's $\alpha$. The former measures agreement, while the latter is based on disagreements, where missing annotations from one or more of the annotators compared do not count towards disagreement \cite{artstein-poesio-2008-survey,bird2009natural}. Results in the \textit{exact} columns require both the start and end character offsets to match across annotators, while \textit{partial} results require only a match on the start offsets. We computed each agreement measure at two levels, namely per span and per character. Since spans might vary in length and include minor differences due to e.g. white-spaces and punctuation marks, character-level scores allow for a more fine-grained comparison. Character-level scores, however, provide somewhat more optimistic estimates, as also all non-annotated characters count towards the agreement measure.

\begin{table}[t]
\caption{\label{tbl:iaa} Inter-annotator agreement for entities.}
\begin{tabular}{lc|cc|cc|cc}
    \toprule
        Type of  &
        Unit of  &
        \multicolumn{2}{c|}{AOA} & \multicolumn{2}{c|}{Fleiss' $\kappa$} &
        \multicolumn{2}{c}{Krippendorff's $\alpha$} \\   
        annotation & agreement & & & & & & \\
        \midrule
         & & Exact  & Partial &  Exact  & Partial & Exact  & Partial\\
         \midrule
	 Entity type & Span & 0.75  & 0.80 & 0.67 & 0.74 & 0.96 & 0.95 \\
	 & Character & 0.96 & -- & 0.86 & -- & 0.94 & -- \\
	 Identifier type & Span & 0.67 & 0.71 & 0.46 & 0.51 & 0.68 & 0.67 \\
	 & Character & 0.94 & -- & 0.79 & -- & 0.69 & --\\
	 Confidential& Span & 0.97 & 0.97 & 0.30 & 0.33  &  0.30 & 0.32 \\
	 attribute & Character  & 0.96 & -- & 0.86 & -- & 0.44 & --\\
\end{tabular}
\end{table}

Agreement on entity types was overall high in terms of partial span-level $\kappa$ and $\alpha$, where automatic pre-annotations likely had a beneficial effect. Exact span-level scores for entities were somewhat lower. This is, however, partly due to minor mismatches (punctuation, spaces) that also count towards disagreement. 
Relatively IAA rates for confidential status were due to the fact that most entities (>96\%) were annotated as not confidential. This dominance of one annotation label leads to a very high expected agreement for $\kappa$ and very low expected disagreement for $\alpha$, which have a negative impact on the resulting IAA scores

For completeness, we also include the identifier type (direct identifier, quasi identifier or no need to mask) in Table \ref{tbl:iaa}. However, it should be stressed that there may be multiple valid sets of masking choices for a given document. Inter-annotator agreement measures are therefore not entirely descriptive for the quality and consistency of the annotation for this label group.

Determining span boundaries was generally rather consistent, likely also due to the existence of pre-annotations. 
A small amount of spans (170) were unusually long, containing $>100$ characters. Most of those long spans correspond were due to abbreviated or translated entity names. 

See the Appendix for more details on disagreements between annotators regarding the entity type and masking decisions. 

\subsection{Use of pre-annotations}

As described in Section \ref{sec:annotation}, the process of creating the initial entities for Step 1 was facilitated by the use of pre-annotations created using a combination of a data-driven NER model with domain-specific heuristics. We inspected how often annotators had to edit those pre-annotations, either by modifying the detected entities, or by inserting new entities that were absent from the pre-annotations. 18.23 \% of all entity mentions were inserted manually by the annotators ($\pm$ 3.25\% depending on the annotator). In total, 23.89 \% of all entity mentions were either inserted or edited by the annotators ($\pm$ 5.14\% depending on the annotator). Annotations for identifier type and confidential attributes were performed manually for every entity as these were not part of pre-annotations.


\subsection{Relations}

As described in Section \ref{sec:relations}, Step 4, annotators were instructed to explicitly mark co-reference relations between nominal entity mentions that were referring to the same underlying entity, but did not have an identical surface realization, such as {\it Government of the Republic of Turkey} and {\it Turkish Government}. Annotators identified in total 5,689 relations, which amount to 3.7\% of all entities. The chance-corrected inter-annotator agreement on relations between mention pairs, as measured by Cohen's $\kappa$, is 0.944. 

Most relations belong to the following entity categories: \ORG{} (3360), \PERSON{} (1,741) and \MISC{} (402). The \PERSON{} co-reference relations largely consisted of various name variants, e.g., \textit{Janice Williams-Johnston} -- \textit{Williams} and titles, e.g., \textit{President of the Court} -- \textit{President}. For the \ORG{} relations there were both a number of organization aliases like \textit{Religious Society of Friends} -- \textit{Quakers}, but also metonymic mentions of country names like \textit{Ireland} -- \textit{Irish Government}. The annotated co-reference relations also encompassed a wide variety of other coreferent expressions, such as money amounts expressed in two currencies (\textit{70,000 Polish zlotys (PLN)} -- \textit{16,400 euros (EUR)}), translations (\textit{Act on Industrial Injury Insurance} -- \textit{Lagen om arbetsskadeförsäkring}) and acronyms along with their definitions (\textit{WBA} -- \textit{Widow's Bereavement Tax Allowance}).

\section{Evaluation metrics}
\label{metrics}

\subsection{Motivation}

Most existing approaches to text anonymization (such as the ones discussed in Section \ref{related}) evaluate their performance by comparing the system's output to human annotations on a held-out dataset. This performance is typically measured using standard IR-based metrics such as {\it precision} and {\it recall}. The recall can be viewed as measuring the degree of \textit{privacy protection}, a high recall representing a low proportion of terms that were left unedited in the document, but which were marked as personal identifiers by human annotators. The precision can be similarly seen as reflecting the remaining \textit{data utility} of the protected documents — the higher the precision, the lower the proportion of terms that were unnecessarily masked, again compared to the choices of human annotators. 

A central objective of data anonymization is to strike a balance between privacy protection and data utility preservation. This balance is often quantified by $F$-scores. However, in the 'privacy-first' approaches to data privacy that underlie most modern approaches to data anonymization \cite{Models}, recall is the most decisive metric. To reflect this precedence of privacy protection over data preservation, several authors have proposed to assign a double weight to the recall, which means using a $F_2$ score instead of the traditional $F_1$ \cite{Ferrandez,Mendels}. Moreover, the use of absolute recall values as a measure of protection/residual risk has been recently brought into question \cite{tkde,acl2021,mozes2021no}, as it relies on a uniform weight over all annotated identifiers and thus fails to account for the fact that some (quasi-)identifiers have a much larger influence on the disclosure risk than others. In particular, failing to detect a direct identifier such as a full person name is much more harmful from a privacy perspective than failing to detect a quasi-identifier.

Another important requirement for the evaluation of anonymization methods is the need to compute the recall at the level of \textit{entities} rather than mentions. Whenever an entity is mentioned several times in a given document, it only makes sense to view this entity as ``protected'' if all of its mentions are masked. If the anonymization method fails to mask at least one mention, the entity will remain readable in the document and will therefore disclose that (quasi-)identifier. For instance, if a person name is mentioned 4 times in a document, and the anonymization method is able to correctly mask three of those mentions, the anonymized text will still retain one mention of that person name in clear text -- an information that can be exploited by an adversary seeking to re-identify the individual we aim to protect. 

Finally, as described in the analysis of inter-annotator agreement in Section \ref{sec:iaa}, text anonymization is a task that may allow for several alternative solutions, as there may be more than one set of masking decisions that satisfy a given privacy property such as $k$-anonymity. In this respect, text anonymization may be likened to other NLP tasks such as conversational agents or machine translation, which also allow for a multiplicity of possible solutions to a given input. The evaluation metrics must therefore be able to compare the system output to multiple expert annotations without presupposing the existence of a unique gold standard. As explained below, this can be achieved by computing recall and precision measures using a \textit{micro-average} over all annotators. 

On the basis of those requirements, we propose to assess the level of protection offered by anonymization methods using a combination of \textit{three evaluation metrics}: 
\begin{itemize}
    \item An \underline{e}ntity-level \underline{r}ecall on \underline{d}irect \underline{i}dentifiers $\textit{ER}_{di}$, 
\item An \underline{e}ntity-level \underline{r}ecall on \underline{q}uasi-\underline{i}dentifiers $\textit{ER}_{qi}$,
\item A token-level \underline{w}eighted \underline{p}recision on both \underline{d}irect and \underline{q}uasi \underline{i}dentifiers $\textit{WP}_{di+qi}$.
\end{itemize}

\subsection{Metrics for privacy protection}

The first two metrics $\textit{ER}_{di}$ and  $\textit{ER}_{qi}$ aim to reflect the degree of privacy protection. Let $D$ denote a set of documents, where each document $d \in D$ is represented as a sequence of tokens. Let $A$ be a set of expert annotators, and $E_{a}(d)$ be the set of entities that were masked by annotator $a$ in the document $d$. Each entity $e \in E_{a}(d)$ is itself defined as a list of token indices $T_e$ where that entity $e$ is mentioned in the document $d$ (there might be several mentions of a given entity in a document). Then, assuming that an anonymization model outputs a set of word indices $M(d)$ to mask in the document $d$, we count each entity $e$ as a true positive if $T_e \subset M(d)$, and a false negative otherwise. In other words, we consider that an entity is correctly masked if and only if the anonymization model manages to completely mask all of its mentions. If that condition is not met, the entity is counted as a false negative.

We use separate recall measures for the direct identifiers (such as full person names, case numbers, etc.) and the quasi-identifiers (dates, locations, etc.). This distinction gives us a more fined-grained measure of the anonymization quality, since a low recall on the direct identifiers corresponds to a failure of the anonymization process (as it implies that the person identity is disclosed), independently of the coverage of other types of identifiers. The set of identifiers $E_{a}(d)$ marked by annotator $a$ in the document $d$ is thus split into two disjoint sets: a set $E_{a}^{di}(d)$ for the direct identifiers and a set $E_{a}^{qi}(d)$ for the quasi-identifiers.

As noted above, a document may admit more than one anonymization solution. To account for this multiplicity, we compute the recall and precision as micro-averages over all annotators. The entity-level recall on direct identifiers $\textit{ER}_{di}$ is thus defined as the micro-averaged recall over the entities defined as direct identifiers:

\begin{equation}
    \textit{ER}_{di} = \frac{\sum_{d \in D} \ \sum_{a \in A} \ \sum_{e \in E_{a}^{di}(d)} \ \mathbf{1}(T_e \subset M(d))} {\sum_{d \in D} \ \sum_{a \in A} \ \left| E_{a}^{di}(d) \right|}
\end{equation}

where $\mathbf{1}$ is the indicator function. An anonymization method will thus obtain a perfect micro-averaged recall if it masks all entities marked as direct identifier by at least one annotator in $A$. The metric implicitly assigns a higher weight to tokens masked by several annotators -- in other words, if all annotators mark a given entity as a direct identifier, not masking it (or not masking it for all of its mentions) will have a larger impact on the recall than an entity masked by a single annotator. 

The entity-level recall on quasi-identifiers $\textit{ER}_{qi}$ is defined similarly:
\begin{equation}
    \textit{ER}_{qi} = \frac{\sum_{d \in D} \ \sum_{a \in A} \ \sum_{e \in E_{a}^{qi}(d)} \ \mathbf{1}(T_e \subset M(d))} {\sum_{d \in D} \ \sum_{a \in A} \ \left| E_{a}^{qi}(d) \right|}
\end{equation}

In order to apply these recall measures, the annotated corpus must satisfy two requirements. First, each text span must be categorized as being either a direct identifier or a quasi-identifier, in order to be able to separately compute $\textit{ER}_{di}$ and $\textit{ER}_{qi}$. Furthermore, the text spans must be grouped into entities. One trivial solution to perform this grouping is to simply cluster mentions that contains the exact same string. However, this strategy fails to account for the multiplicity of referring expressions that may be possible for a given entity (such as ``John Howard'' vs. ``Mr. J. Howard''). In the TAB corpus, this grouping into entities relies on the combination of exact string matching and co-reference relations manually specified by the annotators (see Section \ref{sec:relations}).

\subsection{Metrics for utility preservation}

In addition to offer robust privacy protection, anonymization methods should also maximize the utility of the anonymized outcomes. For text anonymization, this means preserving the semantic content of the protected documents as much as possible. 

As discussed above, the utility preserved by text anonymization methods is usually measured as their {\it precision} in masking sensitive tokens. However, the standard precision metric employed by related works weights the contribution of each unnecessarily masked token uniformly. In contrast, in the PPDP literature on structured data \cite{hundepool2012statistical}, the utility of the anonymized outcomes is measured as the inverse of the information loss resulting from \emph{each} masked element. Therefore, the contribution of masked elements in the remaining data utility will depend on the information they conveyed in the first place, and how this has been affected by the masking operation.

Adapting this notion to text anonymization, we propose weighting the contribution of the masked tokens to the precision according to their informativeness. In information-theoretic terms, the information content $IC$ of a token $t$ is the inverse logarithm of its probability of occurrence: 
$$IC(t) = -log \Pr(t).$$

In this way, rarer (i.e.~more specific) tokens will have higher informativeness than more general/common ones. The notion of $IC$ has been extensively used in the past to measure the semantic content of textual terms \cite{resnik95} and, as stated above, semantics is the most relevant aspect of documents' utility \cite{batet2018semantic}.

$\Pr(t)$ can be estimated in different ways and from a variety of sources \cite{batet20}. In this work we use a masked language model such as BERT \cite{devlin-etal-2019-bert} to estimate $\Pr(t)$ by determining which tokens can be inferred from the rest of the document (as is often the case for e.g.~function words), and which ones represent a more substantial contribution to the document content. The main advantage of BERT w.r.t. a traditional $n$-gram model is the possibility to exploit a much broader document context to compute the probabilities, and therefore the information content of each span.

Concretely, we compute this probability by applying BERT (with a language modeling head on top) on the full document where all masked tokens are replaced by the special \ENT{[MASK]} symbol. Then, one can then define the information content $IC(t,d)$ of the masked token $t$ at position $i$ in document $d$ as:
\begin{equation}
IC(t,d) = - \log(BERT(d)[i,t])
\end{equation}
where $BERT(d)[i,t]$ denotes the probability value of token $t$ predicted by BERT at position $i$ in the document. A high probability value reflects that the token is predictable from the document context, and therefore has a low information content. In contrast, a low probability value indicates a token that cannot be inferred from the remaining part of the document, and has therefore a high information content. 

As for the aforementioned metrics for privacy protection, measures of utility preservation should also account for the fact that the anonymization of a given document may lead to several equally valid solutions. This can be similarly expressed through a micro-average over annotators. Putting it all together, we define the weighted token-level precision on all identifiers $\textit{WP}_{di+qi}$ as 
\begin{equation}
    \textit{WP}_{di+qi} = \frac{\sum_{d \in D} \ \sum_{a \in A} \ \sum_{t \in M(d)} \ \mathbf{1}(t \in T_a(d)) \cdot IC(t,d)} {|A| \sum_{d \in D} \ \sum_{t \in M(d)} \ IC(t,d) }
\end{equation}

where $T_a(d) = \bigcup_{e \in E_a(d)} \ T_e$ represents all tokens masked by annotator $a$ in document $d$, including both direct identifiers and quasi-identifiers. In other words, a low precision indicates that the anonymization model has masked many tokens that, according to the expert annotator(s), did not need to be concealed. Since each token is weighted by its information content $IC(t,d)$, tokens that convey more information (or, equivalently, are more difficult to predict from the edited document) lead to a higher penalty on the precision measure. 

\subsection{Example}

Table \ref{table:example} illustrates an example of a short paragraph including two human annotators and the output of two anonymization models. 

\begin{table}[t]
\caption{\label{table:example}Short paragraph annotated by two human annotators. Each annotated text span is associated with an entity ID and a type (direct or quasi). Below the human annotations, we also show the masking decisions of two anonymization models, Longformer and Presidio (cf. next section).}
\begin{framed}
\includegraphics[scale=0.38]{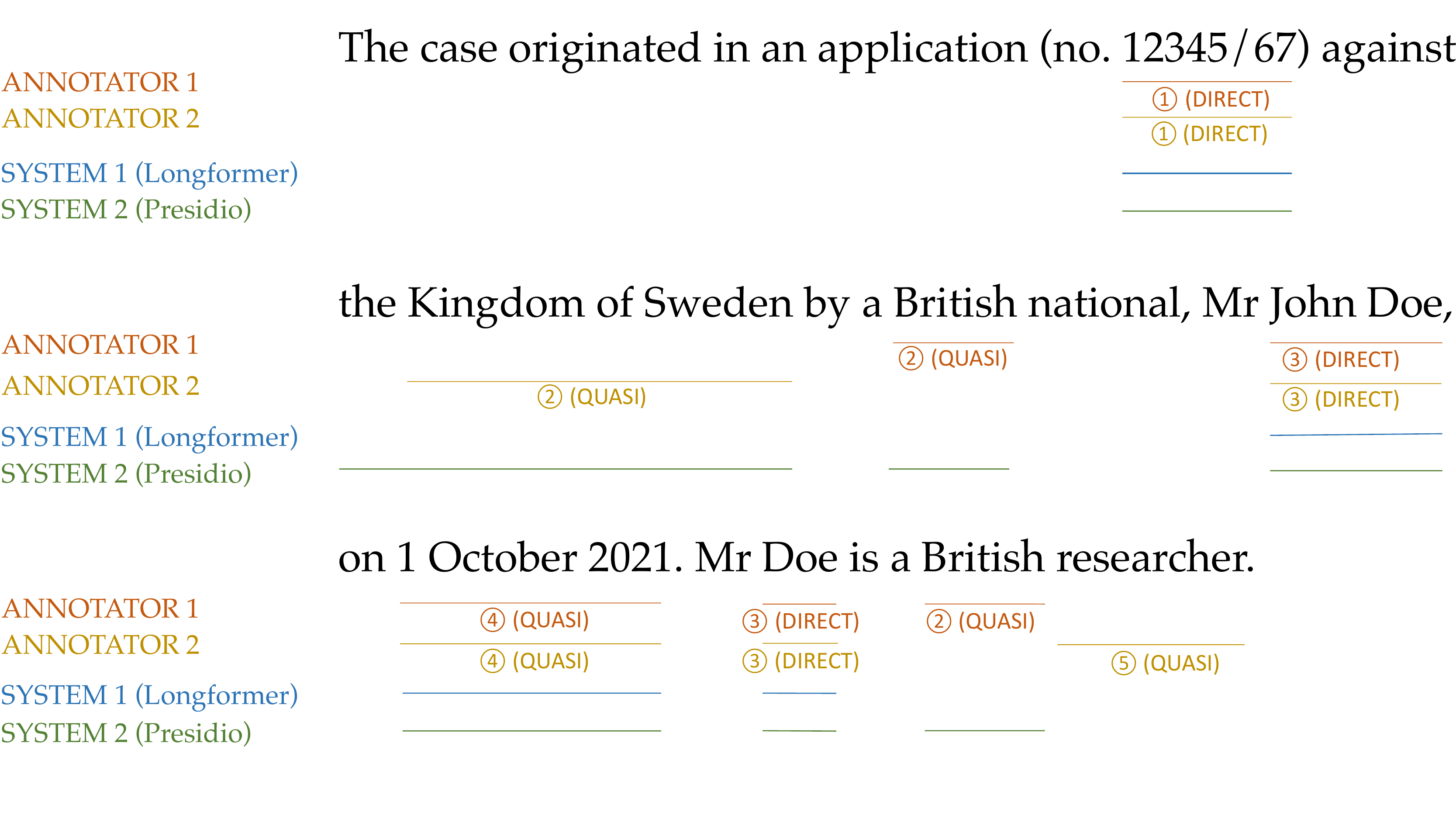}
\vspace{-6mm}
\end{framed}
\end{table}

We can observe in this example that both annotators have marked two direct identifiers: \textit{no. 12345/67} and  \textit{John Doe / Doe}. The annotators have, however, decided to select different quasi-identifiers: while the first annotator marked \textit{British} and \textit{1 October 2021}, the second annotator decided to mask \textit{Kingdom of Sweden}, \textit{1 October 2021}, and \textit{researcher}. 

The first anonymization model (Longformer) correctly identified all direct identifiers, resulting in $\textit{ER}_{di}=\frac{2+2}{2+2}=1$. However, the model has only masked the date as quasi identifier, giving $\textit{ER}_{qi}=\frac{1+1}{2+3}=0.4$. Since all 7 tokens belonging to the four entity mentions masked by the model were also masked by both annotators, $\textit{WP}_{di+qi}=\frac{7+7}{7+7}=1$ if we assume for simplicity that all tokens have a uniform $IC$ value set to 1. 

The second anonymization tool (Presidio) detects the person's name but leaves the case number unmasked, yielding $\textit{ER}_{di}=\frac{2+0}{2+2}=0.5$. The tool also detects 
the quasi-identifier \textit{Kingdom of Sweden} (marked by the second annotator) and the date (marked by both annotators), resulting in $\textit{ER}_{qi}=\frac{1+2}{2+3}=0.6$. The token-level precision of this second model -- assuming again uniform $IC$ weights for all tokens -- is $\textit{WP}_{di+qi}=\frac{9+10}{13+13}\approx 0.73$ since out of a total of 13 tokens masked by the tool, 9 and 10 tokens were also respectively masked by the first and second annotator.

\section{Empirical results}
\label{results}

To illustrate the use and potential results derived from our evaluation framework, in this section we report empirical results on the evaluation of three well-differentiated systems for text anonymization. The first two systems correspond to baselines that rely on existing neural models for Named Entity Recognition (NER), while the third approach is explicitly fine-tuned on the masking decisions of the TAB corpus.

\subsection{Baseline Performance}
\label{baseline}

The first anonymization model relies on a neural named entity recognition model based on the RoBERTa language model \citep{DBLP:journals/corr/abs-1907-11692} and fine-tuned for NER on the Ontonotes v5 \citep{ontonotes2011}, as implemented in spaCy \cite{spacy}. The anonymization masked the full set of 18 categories from Ontonotes (\ENT{PERSON}, \ENT{ORG}, \ENT{GPE},  \ENT{LOC}, \ENT{DATE}, \ENT{TIME}, \ENT{LANGUAGE}, \ENT{CARDINAL},  \ENT{EVENT}, \ENT{FAC}, \ENT{LAW}, \ENT{LOC}, \ENT{MONEY}, \ENT{NORP}, \ENT{ORDINAL}, \ENT{PERCENT}, \ENT{PRODUCT}, \ENT{QUANTITY}, \ENT{WORK\_OF\_ART}), with the exception of \ENT{CARDINAL} entities where only occurrences comprising at least four digits were masked to avoid too many false positives.

The second anonymization tool is Presidio\footnote{Version 2.2.23. \url{https://github.com/microsoft/presidio}}, a data protection \& anonymization API developed by Microsoft that relies on a combination of template-based and NER-based machine learning models to detect and mask personally identifiable information in text. Compared to the generic neural NER model above, Presidio's named entity types, models and rules are explicitly targeted towards data privacy. The tool masked the following entity types: \ENT{PERSON}, \ENT{LOCATION}, \ENT{NRP} (nationality, religious or political group), \ENT{DATE\_TIME}, contact information (email address, phone number) and various codes and numbers (driver license, bank account, identification numbers, etc.). We provide evaluation results for Presidio under two configuration settings, namely the default mode and one in which the detection of organization names (governments, public administration, companies, etc.) is also activated. 

Table \ref{tbl:NER} reports the evaluation results for both systems on the development and test sets of our corpus. In particular we report the standard {\it precision} and {\it recall} metrics employed in related works (which give uniform weights to all terms and assess matches at the level of entity mentions), and the new privacy and utility metrics we presented in Section \ref{metrics}.

\begin{table}[h]
\caption{\label{tbl:NER} Evaluation results for the two baselines (generic neural NER model and Microsoft Presidio) on the development and test sections of the TAB corpus. We report both the standard, token-level recall $R_{di+qi}$ and precision $P_{di+qi}$ on all identifiers (micro-averaged over all annotators) as well as the three proposed evaluation metrics $\textit{ER}_{di}$, $\textit{ER}_{qi}$ and $\textit{WP}_{di+qi}$ from Section \ref{metrics}.} 
\begin{tabular}{llccccc}  
	\toprule  
	  System & Set & $R_{di+qi}$ & $\textit{ER}_{di}$ & $\textit{ER}_{qi}$ & $P_{di+qi}$ & $\textit{WP}_{di+qi} $  \\
	  \midrule 
Neural NER (RoBERTa   &	Dev & 0.910  & 0.970 & 0.874 & 0.447 & 0.531 \\ 
fine-tuned on Ontonotes v5)            &   Test & 0.906  & 0.940 & 0.874 & 0.441 &  0.515\\ 
	\midrule 
 Presidio (default) &  Dev & 0.696 & 0.452 & 0.739 & 0.771 & 0.795  \\ 
                    & Test & 0.707 & 0.460 & 0.758 & 0.761 & 0.790 \\ 
    \midrule 
Presidio (+ORG) & Dev & 0.767 & 0.465 & 0.779 & 0.549 & 0.622 \\ 
                & Test & 0.782 & 0.463 & 0.802 & 0.542 & 0.609  \\
\end{tabular}
\end{table}

Presidio's results illustrate the importance of computing separate recall measures for the direct and the quasi identifiers: although the standard, mention-level recall seems relatively good at first sight (around 0.7), a closer look a the entity-level recall over direct identifiers $\textit{ER}_{di}$ shows a much poorer performance (around 0.45). Since quasi-identifiers are typically much more frequent than direct identifiers, this poor performance on direct identifiers (which are the most harmful entities from a privacy perspective) is easy to miss if one conflates all identifiers in a single recall measure. In particular, Presidio fails to detect all court case numbers (a category it was not trained on), which is a publicly known information that unequivocally identifies the case applicant. Even though Presidio was able to detect other direct identifiers such as the applicant's name, failing to detect case numbers will likely render the anonymization useless. 
In contrast, the generic NER model got significantly better results w.r.t. direct identifiers because, coincidentally, case numbers matched the generic \CARDINAL{} class from Ontonotes. However, this would no longer hold if case numbers had been alphanumeric rather than numeric. This shows that not only quasi-identifiers, but also direct identifiers cannot be limited to predefined categories of entities.

On the other hand, the generic NER model performed poorly w.r.t. precision: due to the large variety of general categories it considers, more than half of the tagged entities dit not need to be masked. A similar behavior can be observed for Presidio when enabling the quite general ORG category: recall sightly increases at the cost of a significantly lower precision. Indeed, the goal of the anonymization process is not to detect all occurrences of predefined semantic categories, but to mask only entities that refer to the individual to be protected. Court cases are quite rich in contextual information that do not refer to the applicant or the application, but to laws or procedures. As a result, a substantial number of persons, places and organizations do not need to be edited out. This illustrates another limitation of NER-based systems, namely the fact that they incur in unnecessary masking that hampers the utility and readability of the protected outcomes. In summary, both systems offered a poor balance between recall/protection and precision/utility preservation.

We also observe that the weighted utility metric we propose in Section \ref{metrics} ($\textit{WP}_{di+qi}$) results in higher figures than the standard precision. Considering that our metric weights the contribution of each unnecessarily masked term according to its information content (whereas the standard precision treats them uniformly), this indicates that these terms are less informative (that is, more general) than the average. This is also consistent with the fact that these unmasked terms do not seem to incur in privacy disclosure according to the annotators, either because they are not related to the individual to be protected or due to they provide very general information. Therefore, our metric provides a more accurate estimation of the utility preserved in the protected output than the standard precision. 



\subsection{Performance of fine-tuned models}
\label{best}

Contrary to the first two baseline systems, which are zero-shot approaches that did not rely on any form of domain adaptation, the third approach is explicitly fine-tuned on the training set of the TAB corpus. More specifically, we evaluate the anonymization performance by fine-tuning a large pretrained language model, the Longformer \citep{beltagy2020longformer} model, which is a BERT-style model built from the RoBERTa \cite{liu2019roberta} checkpoint and pretrained for masked language modelling on long documents. It uses a modified attention mechanism, namely a combination of a sliding window (local) attention and global attention,  which allows for processing of longer documents. This addresses the downside of BERT-style models which are able to process up to 512 tokens, and for which existing techniques either shorten the text or partition it, thereby resulting in a loss of contextual information. Early experiments with a RoBERTa model have indeed shown a negative difference of 2-4 percentage points in $F_1$ score compared to the Longformer model.

The LongFormer model \footnote{\url{https://huggingface.co/allenai/longformer-base-4096}} was fine-tuned on the training set of the TAB corpus with a linear inference layer on top to predict which text span should be masked using IOB sequence labelling. When multiple annotations were available for the same court case, we chose to treat all annotations as equally correct and duplicate the document for each distinct annotation layer. Each token received a \ENT{MASK} label if its identifier type was set by the annotator to either \DIRECT{} or \QUASI{}. 


The experimental results obtained with various settings of this LongFormer model are shown in Table \ref{tbl:hyperparameters}. We provide results for different window sizes (the maximum allowed by LongFormer being 4096 tokens). Furthermore, to reflect the higher importance attached to the recall in text anonymization, we also provide results for various label weights set on the cross-entropy loss.\footnote{In other words, a label weight of e.g.~(5,1) stipulates that the cost of a false negative (omitting to mask a token that should have been masked) is five times the cost of a false positive.} The optimization used the AdamW optimizer \cite{loshchilov2018decoupled} with a learning rate of $2 \times 10^{-5}$ and 2 epochs. The above hyper-parameters were all selected experimentally on the development set. 


\begin{table}[h]
\caption{\label{tbl:hyperparameters} Experimental results for the fine-tuned LongFormer model on the development set depending on the context window size and the label weight.} 
\begin{tabular}{lllccccc}  
	\toprule  
	  Window size & Label weight & $R_{di+qi}$ & $\textit{ER}_{di}$ &
	  $\textit{ER}_{qi}$ & $P_{di+qi}$ & $\textit{WP}_{di+qi} $  \\
	  (nb tokens) & \scriptsize{(MASK, NO\_MASK)} & & & & & \\
\midrule 
32 & (1,1)   & 0.875 & 0.987 & 0.850 & 0.612 & 0.643 \\ 
32 & (5,1)   & 0.924 & 0.992 & 0.915 & 0.550 &  0.583\\ 
32 & (10,1)   & 0.954 & 0.994 & 0.952 & 0.470 &  0.507\\ 
\midrule 
512 & (1,1)   & 0.847 & 0.986 & 0.817 & \textbf{0.929} & \textbf{ 0.932}\\ 
512 & (5,1)   & 0.914 & 0.993 & 0.906 & 0.858 & 0.866 \\ 
512 & (10,1)   & \textbf{0.937} & \textbf{0.997} & 0.930 & 0.767 & 0.783\\ 
\midrule 
4096 & (1,1) & 0.860 & 0.988 & 0.847 & 0.925 & 0.929 \\ 
4096 & (5,1) & 0.916 & 0.988 & 0.913 & 0.843 &  0.856\\ 
4096 & (10,1) & 0.935 & 0.993 & \textbf{0.936} & 0.795 &  0.811\\ 
	\midrule 
\end{tabular}
\end{table}

Unsurprisingly, the fine-tuned LongFormer model outperforms the previous baselines from Table \ref{tbl:NER}. As expected, we also notice that the performance on the recall metrics improve for higher weights associated to the \textit{MASK} label, while a larger window size makes it possible to take into account a broader context, and therefore a higher precision. We select the Longformer model with the maximum window size and label weight of (10,1) for the final evaluation, shown in Table \ref{tbl:all}. 

\begin{table}[h]
\caption{\label{tbl:all} Evaluation result for the fine-tuned LongFormer model with a window size of 4096 and a label weight of (10,1) on the development and test section of the TAB corpus.} 
\begin{tabular}{lccccc}  
	\toprule  
	  Set & $R_{di+qi}$ & $\textit{ER}_{di}$ & $\textit{ER}_{qi}$ & $P_{di+qi}$ & $\textit{WP}_{di+qi} $  \\
\midrule 
 Dev &	0.935 & 0.993 & 0.936 & 0.795 &  0.811\\
Test & 0.919 & 1.000 & 0.916 & 0.836 &  0.850\\ 
	\midrule 
\end{tabular}
\end{table}

\begin{figure}[t]
    \centering
        \centering
        \caption{\label{test}Proportion of false negatives factored per entity type in the development and test sets. The relative frequency of each type in the corpus is given in parenthesis next to the entity type.\\}
        \includegraphics[width=\linewidth]{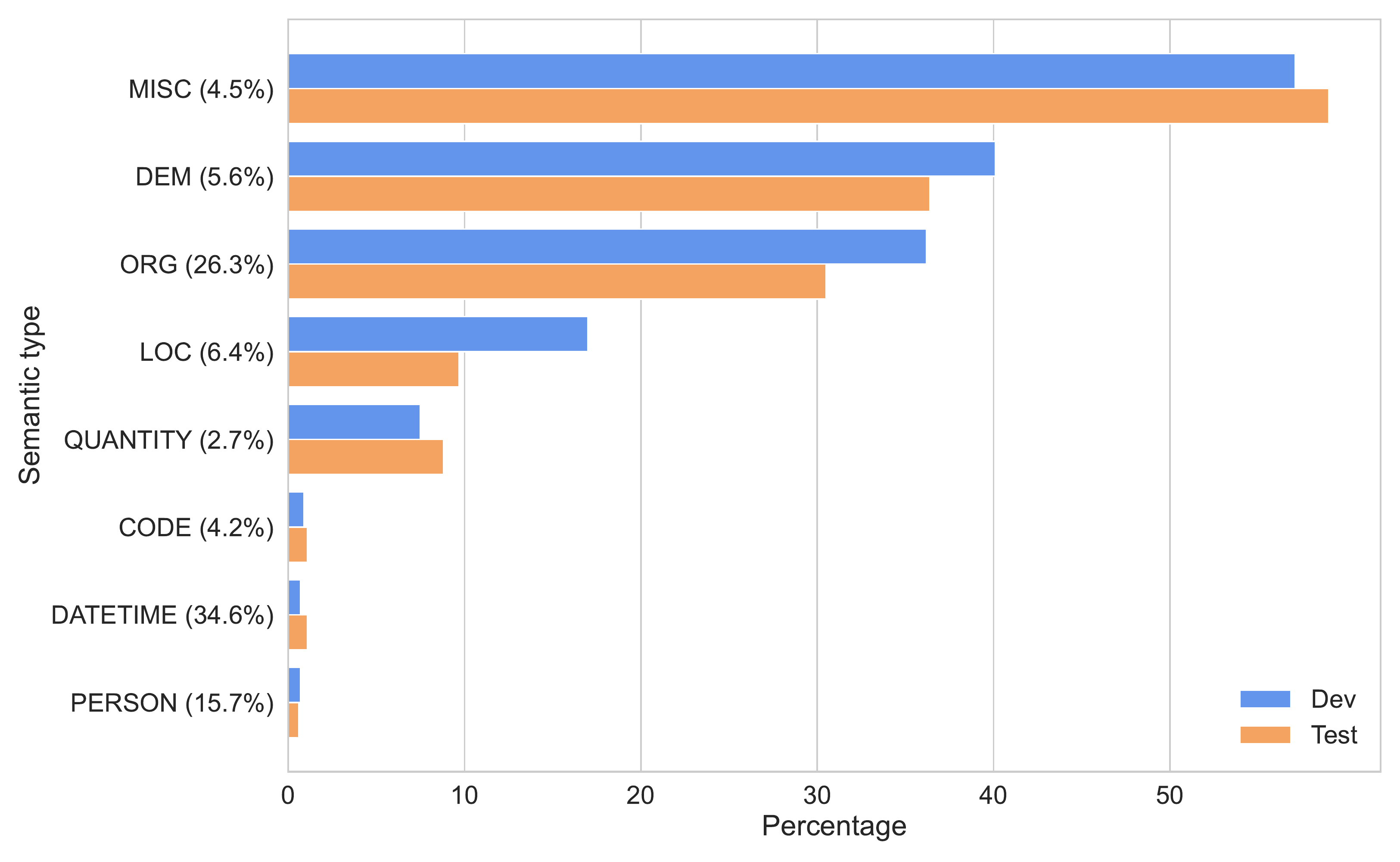}
\end{figure}

Figure \ref{test} shows the distribution of false negatives per semantic type (see \ref{semantic types}), for both the development and the test set. We notice that the entity types \MISC{}, \DEM{} and \ORG{} are substantially more difficult to handle that the other entity types. For \MISC{} and \DEM{}, this difficulty seems to stem primarily from the difficulty to detect the occurrences of those entities, as, in contrast to the other entity types, they do not correspond to named entities and may take a broad variety of forms. For \ORG{}, the proportion of false negatives stems from the fact that masking decisions related to this type of organizations seem to be particularly difficult -- even for human annotators, as evidenced by the relatively large proportion of \QUASI{} vs. \NOMASK{} disagreements for this entity type (see Table \ref{tbl:masking_disagr_per_entity}). This problem is particularly salient for the mentions of regional courts, which are often mentioned in ECHR court cases, and which often provide indirect cues about the place of residence of the applicant. We also provide in Table \ref{table:examples} a few examples of masking errors, categorized in four distinct error types. 


\begin{table}[t!]
\caption{\label{table:examples}Examples of masking errors. The true span according to the human annotators is compared against the model's prediction.}
\begin{framed}
\includegraphics[scale=0.45]{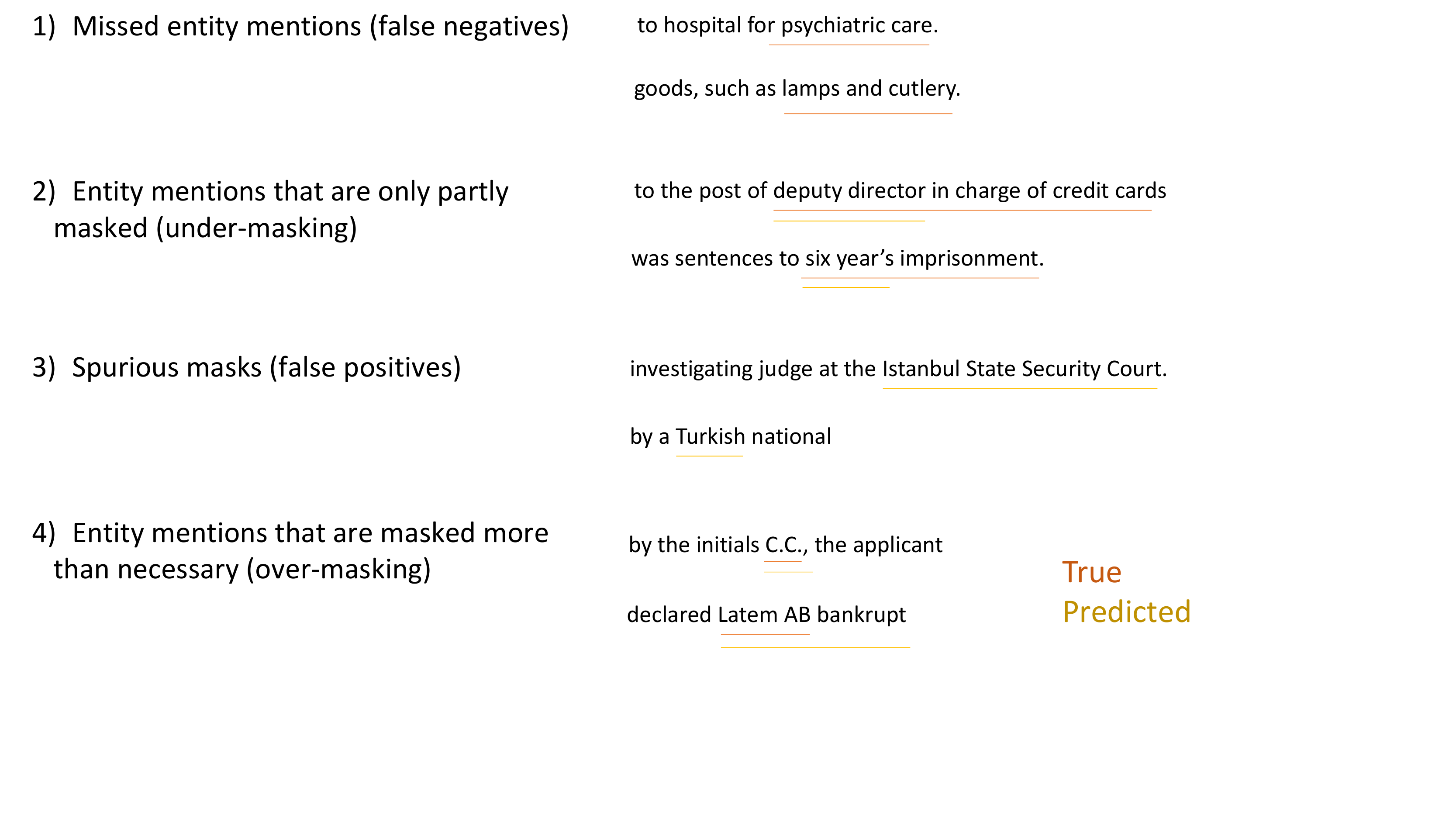}
\end{framed}
\end{table}
\clearpage

\subsection{Performance on out-of-domain data}

To assess whether the fined-tuned anonymization model could be applied on out-of-domain data (and more specifically on texts outside of the legal domain), we also evaluate its performance on a dataset of Wikipedia biographies that we have manually annotated for text anonymization \cite{domainspecific}. Biographies are rich in personal information of a more general nature since they describe a large number of different individuals. 

This dataset, which was constructed as a follow-up to the TAB corpus, consists of 553 biographies, 20 of which were annotated by more than one annotator. The annotation process and guidelines for this dataset were very similar to the process described in Section \ref{corpus}, with each text span being assigned a semantic type and a identifier type. The only substantial difference in the two annotation processes was the absence of annotation for the confidential status of each entity. The dataset is, however, substantially smaller than the TAB corpus, with about 11,217 annotated entities, compared to a total of 108,151 entities for TAB. 

Table \ref{tbl:domain} presents the evaluation results of two anonymization models when applied to this dataset of Wikipedia biographies. The first is the neural NER model presented in Section \ref{baseline}, which consists of a RoBERTa language model fine-tuned for Named Entity Recognition on Ontonotes v5. The second model is the Longformer model from Section \ref{best}, fine-tuned on the training set of TAB. 

\begin{table}[h!]
\caption{\label{tbl:domain} Evaluation results for the fine-tuned LongFormer model on the manually annotated Wikipedia dataset. A generic neural NER model was also used for comparison.} 
\begin{tabular}{lccccc}  
	\toprule  
	   & $R_{di+qi}$ & $\textit{ER}_{di}$ & $\textit{ER}_{qi}$ & $P_{di+qi}$ & $\textit{WP}_{di+qi} $  \\
\midrule 
 RoBERTa (fined-tuned on Ontonotes) &	0.845 & 0.810 & 0.801 & 0.770 &  0.836\\
 \midrule
Longformer (fine-tuned on TAB) & 0.952 & 0.999 & 0.923 & 0.708 &  0.765\\ 
	\midrule 
\end{tabular}
\end{table}

As we can observe in Table \ref{tbl:domain}, the Longformer model fine-tuned on the training set of TAB can detect a relatively large number of personal information in the Wikipedia biographies, despite the substantial differences in the linguistic form of those texts. In comparison, the generic RoBERTa model trained on the named entities from Ontonotes misses a sizable number of both direct and indirect identifiers. 



\section{Conclusions and future work}
\label{conclusion}

We have presented a novel benchmark and associated evaluation metrics for text anonymization evaluation. Compared to the evaluation corpora available in the literature, the TAB corpus is larger (accounting more than a thousand documents), more general (due to the broad range of biographical details mentioned in court cases), freely available in clear form (due to not being subjected to privacy issues), and explicitly targeted towards text anonymization. 

Manual annotation efforts are inherently limited by the presence of residual errors, omissions, inconsistencies, or differences in human judgments. Human annotations cannot provide any formal privacy guarantees, in contrast to methods based on explicit privacy models such as $k$-anonymity and its extensions \cite{Samarati,Samarati_Sweeney:1998,Li:2007} or differential privacy \citep{Dwork2006a,10.1561/0400000042}. However, those issues were mitigated in the development of the TAB corpus by:
\begin{itemize}
    \item defining domain-independent and privacy/GDPR-oriented annotation guidelines, and avoiding in particular to restrict the anonymization process to predefined lists of entity types;  
\item aggregating the masking decisions of several human annotators in the evaluation metrics using micro-averages. 
\end{itemize}

The annotators were instructed to determine which the entities to mask based on “publicly available knowledge”, for instance through information that can be gathered on the web. The annotations are therefore dependent on the assumption that potential attackers do not have access to other (non-public) sources of background knowledge related to the court cases or the individuals mentioned in them. The need to rely on an explicit assumption regarding the available background knowledge is another important difference between the annotation strategy presented in this paper and approaches based on differential privacy, which do not need to rely on such assumptions. 

The Text Anonymization Benchmark aims to facilitate the evaluation and comparison of anonymization algorithms for textual data, and provide a more accurate assessment of the actual privacy protection they achieve. In particular, we show in Section \ref{results} that NER-based methods, which constitute the \textit{de facto} approach for text anonymization in most domains (to the possible exception of biomedical texts), offer weak protection. We also have shown how a baseline model based on a pre-trained language model explicitly fine-tuned for anonymization is able to provide  stronger protection against disclosure and a better balance between privacy and data utility preservation.

In addition to the TAB corpus, we have also presented a set of evaluation metrics for text anonymization that go beyond the standard IR-based metrics employed in the literature. The novelty of our metrics consists of weighting the contribution of each term according to the level of disclosure risk it incurs and the information it conveys. With this we provide a more comprehensive and accurate assessment of the performance of anonymization methods, as we have shown in the empirical experiments.

The Text Anonymization Benchmark focuses on detecting and annotating terms that may cause disclosure, and must be subsequently concealed or masked. Nevertheless, since the best anonymization is such that it optimizes the trade-off between privacy protection and data utility preservation, replacing sensitive terms by less detailed versions (i.e., generalizations) would be preferred to just suppressing them. Current de-identification/anonymization methods are typically limited to term suppression or, at most, to replacing sensitive terms by their semantic categories (such as replacing "John Doe" with "[\PERSON{}]"). However, some utility-preserving generalization-based methods have been proposed \cite{tPlaus,Sanchez2016,Sanchez2017,Ksafety}. As future work, we plan to extend our benchmark by incorporating privacy-preserving replacements for masked terms. This is certainly challenging, because multiple combinations of replacements could be equally valid to prevent disclosure, but only one would be optimal from the perspective of data utility preservation. This selection of privacy-preserving replacements will enable us to better evaluate the loss of utility incurred by anonymization methods, and provide a ground truth for optimal utility-preserving masking. We also expect that our utility metric will be of great help to accurately quantify the utility loss.
%


\begin{acknowledgments}
We acknowledge support from the Norwegian Research Council (CLEANUP project (\url{http://cleanup.nr.no/}), grant nr. 308904) and the Government of Catalonia (ICREA Acadèmia Prize to D. Sánchez). We also wish to thank the 12 law students who contributed to the annotation process: Isak Falch Alsos, Saba Abadhar, Sigurd Teofanovic, Vilde Katrin Lervik, Sarah Kristin Geisler, Louise Øverås Nilsen, Marlena Zaczek, Ole Martin Moen, Nina Stærnes, Rose Monrad, Selina Ovat and Alexandra Kleinitz Schultz. The views in this paper are not necessarily shared by UNESCO.
\end{acknowledgments}

\vspace{2cm} 

\starttwocolumn
\bibliography{biblio}

\begin{thebibliography}{96}
\expandafter\ifx\csname natexlab\endcsname\relax\def\natexlab#1{#1}\fi

\bibitem[{Aberdeen et~al.(2010)Aberdeen, Bayer, Yeniterzi, Wellner, Clark,
  Hanauer, Malin, and Hirschman}]{aberdeen2010mitre}
Aberdeen, John, Samuel Bayer, Reyyan Yeniterzi, Ben Wellner, Cheryl Clark,
  David Hanauer, Bradley Malin, and Lynette Hirschman. 2010.
\newblock The {MITRE} identification scrubber toolkit: design, training, and
  assessment.
\newblock \emph{International Journal of Medical Informatics}, 79(12):849--859.

\bibitem[{Alfalahi, Brissman, and Dalianis(2012)}]{Alf:Bri:Dal:2012}
Alfalahi, Alyaa, Sara Brissman, and Hercules Dalianis. 2012.
\newblock Pseudonymisation of personal names and other {PHI}s in an annotated
  clinical {S}wedish corpus.
\newblock In \emph{Third LREC Workshop on Building and Evaluating Resources for
  Biomedical Text Mining (BioTxtM 2012)}, pages 49--54.

\bibitem[{Anandan and Clifton(2011)}]{Relationships}
Anandan, Balamurugan and Chris Clifton. 2011.
\newblock Significance of term relationships on anonymization.
\newblock In \emph{Proceedings of the 2011 IEEE/WIC/ACM International Joint
  Conference on Web Intelligence and Intelligent Agent Technology - Workshops,
  WI-IAT 2011}, pages 253--256, Lyon, France.

\bibitem[{Anandan et~al.(2012)Anandan, Clifton, Jiang, Murugesan,
  Pastrana-Camacho, and Si}]{tPlaus}
Anandan, Balamurugan, Chris Clifton, Wei Jiang, Mummoorthy Murugesan, Pedro
  Pastrana-Camacho, and Luo Si. 2012.
\newblock $t$-plausibility: Generalizing words to desensitize text.
\newblock \emph{Transactions on Data Privacy}, 5(3):505--534.

\bibitem[{Artstein and Poesio(2008)}]{artstein-poesio-2008-survey}
Artstein, Ron and Massimo Poesio. 2008.
\newblock Survey article: Inter-coder agreement for computational linguistics.
\newblock \emph{Computational Linguistics}, 34(4):555--596.

\bibitem[{Barrett et~al.(2019)Barrett, Kementchedjhieva, Elazar, Elliott, and
  S{\o}gaard}]{barrett-etal-2019-adversarial}
Barrett, Maria, Yova Kementchedjhieva, Yanai Elazar, Desmond Elliott, and
  Anders S{\o}gaard. 2019.
\newblock Adversarial removal of demographic attributes revisited.
\newblock In \emph{Proceedings of the 2019 Conference on Empirical Methods in
  Natural Language Processing and the 9th International Joint Conference on
  Natural Language Processing (EMNLP-IJCNLP)}, pages 6330--6335, Association
  for Computational Linguistics, Hong Kong, China.

\bibitem[{Batet and S{\'a}nchez(2018)}]{batet2018semantic}
Batet, Montserrat and David S{\'a}nchez. 2018.
\newblock Semantic disclosure control: semantics meets data privacy.
\newblock \emph{Online Information Review}, 42(3):290--303.

\bibitem[{Batet and S{\'a}nchez(2020)}]{batet20}
Batet, Montserrat and David S{\'a}nchez. 2020.
\newblock Leveraging synonymy and polysemy to improve semantic similarity
  assessments based on intrinsic information content.
\newblock \emph{Artificial Intelligence Review}, 53(3):2023--2041.

\bibitem[{Beltagy, Lo, and Cohan(2019)}]{beltagy2019scibert}
Beltagy, Iz, Kyle Lo, and Arman Cohan. 2019.
\newblock Scibert: A pretrained language model for scientific text.
\newblock \emph{arXiv preprint arXiv:1903.10676}.

\bibitem[{Beltagy, Peters, and Cohan(2020)}]{beltagy2020longformer}
Beltagy, Iz, Matthew~E. Peters, and Arman Cohan. 2020.
\newblock Longformer: The long-document transformer.

\bibitem[{Bick and Barreiro(2015)}]{bick2015}
Bick, Eckhard and Anabela Barreiro. 2015.
\newblock Automatic anonymisation of a new portuguese-english parallel corpus
  in the legal-financial domains.
\newblock \emph{Oslo Studies in Language}, 7(1):101--124.

\bibitem[{Bier et~al.(2009)Bier, Chow, Golle, King, and Staddon}]{Bier}
Bier, Eric~A., Richard Chow, Philippe Golle, Tracy~H. King, and J.~Staddon.
  2009.
\newblock The rules of redaction: Identify, protect, review (and repeat).
\newblock \emph{IEEE Security and Privacy Magazine}, 7(6):46--53.

\bibitem[{Bird, Klein, and Loper(2009)}]{bird2009natural}
Bird, Steven, Ewan Klein, and Edward Loper. 2009.
\newblock \emph{Natural language processing with Python: analyzing text with
  the natural language toolkit}.
\newblock " O'Reilly Media, Inc.".

\bibitem[{Blodgett, Green, and O{'}Connor(2016)}]{blodgett2016demographic}
Blodgett, Su~Lin, Lisa Green, and Brendan O{'}Connor. 2016.
\newblock Demographic dialectal variation in social media: A case study of
  {A}frican-{A}merican {E}nglish.
\newblock In \emph{Proceedings of the 2016 Conference on Empirical Methods in
  Natural Language Processing}, pages 1119--1130, Association for Computational
  Linguistics, Austin, Texas.

\bibitem[{Boisen et~al.(2000)Boisen, Crystal, Schwartz, Stone, and
  Weischedel}]{boisen-etal-2000-annotating}
Boisen, Sean, Michael~R. Crystal, Richard Schwartz, Rebecca Stone, and Ralph
  Weischedel. 2000.
\newblock Annotating resources for information extraction.
\newblock In \emph{Proceedings of the Second International Conference on
  Language Resources and Evaluation ({LREC}{'}00)}, European Language Resources
  Association (ELRA), Athens, Greece.

\bibitem[{Bommasani et~al.(2019)Bommasani, Wu, Zhiwei, and
  Schofield}]{neurips2019_synthesis}
Bommasani, Rishi, Steven Wu, Zhiwei, and Alexandra~K Schofield. 2019.
\newblock Towards private synthetic text generation.
\newblock In \emph{NeurIPS 2019 Workshop on Machine Learning with Guarantees},
  Vancouver, Canada.

\bibitem[{Chakaravarthy et~al.(2008)Chakaravarthy, Gupta, Roy, and
  Mohania}]{Ksafety}
Chakaravarthy, Venkatesan~T., Himanshu Gupta, Prasan Roy, and Mukesh~K.
  Mohania. 2008.
\newblock Efficient techniques for document sanitization.
\newblock In \emph{Proceedings of the 17th ACM Conference on Information and
  Knowledge Management, CIKM 2008}, pages 843--852, Napa Valley, California,
  USA.

\bibitem[{Chiu and Nichols(2016)}]{Chi:Nic:16}
Chiu, Jason~P.C. and Eric Nichols. 2016.
\newblock Named entity recognition with bidirectional {LSTM}-{CNN}s.
\newblock \emph{Transactions of the Association for Computational Linguistics},
  4:357--370.

\bibitem[{Chow, Golle, and Staddon(2008)}]{Chow}
Chow, Richard, Philippe Golle, and Jessica Staddon. 2008.
\newblock Detecting privacy leaks using corpus-based association rules.
\newblock In \emph{Proceedings of the 14th ACM SIGKDD International Conference
  on Knowledge Discovery and Data Mining}, KDD '08, page 893–901, Association
  for Computing Machinery, New York, NY, USA.

\bibitem[{Cohen(2012)}]{cohen2012privacy}
Cohen, Julie~E. 2012.
\newblock What privacy is for.
\newblock \emph{Harvard Law Review}, 126:1904.

\bibitem[{Cumby and Ghani(2011)}]{Kconfu}
Cumby, Chad~M. and Rayid Ghani. 2011.
\newblock A machine learning based system for semi-automatically redacting
  documents.
\newblock In \emph{Proceedings of the Twenty-Third Conference on Innovative
  Applications of Artificial Intelligence}, pages 1628--1635, San Francisco,
  California, USA.

\bibitem[{Dernoncourt et~al.(2017)Dernoncourt, Lee, Uzuner, and
  Szolovits}]{dernoncourt2017identification}
Dernoncourt, Franck, Ji~Young Lee, Ozlem Uzuner, and Peter Szolovits. 2017.
\newblock De-identification of patient notes with recurrent neural networks.
\newblock \emph{Journal of the American Medical Informatics Association},
  24(3):596--606.

\bibitem[{Devlin et~al.(2018)Devlin, Chang, Lee, and
  Toutanova}]{devlin2018bert}
Devlin, Jacob, Ming-Wei Chang, Kenton Lee, and Kristina Toutanova. 2018.
\newblock {BERT}: Pre-training of deep bidirectional transformers for language
  understanding.
\newblock \emph{arXiv preprint arXiv:1810.04805}.

\bibitem[{Devlin et~al.(2019)Devlin, Chang, Lee, and
  Toutanova}]{devlin-etal-2019-bert}
Devlin, Jacob, Ming-Wei Chang, Kenton Lee, and Kristina Toutanova. 2019.
\newblock {BERT}: Pre-training of deep bidirectional transformers for language
  understanding.
\newblock In \emph{Proceedings of the 2019 Conference of the North {A}merican
  Chapter of the Association for Computational Linguistics: Human Language
  Technologies, Volume 1 (Long and Short Papers)}, pages 4171--4186,
  Association for Computational Linguistics, Minneapolis, Minnesota.

\bibitem[{Domingo-Ferrer, S{\'a}nchez, and Soria-Comas(2016)}]{Models}
Domingo-Ferrer, Josep, David S{\'a}nchez, and Jordi Soria-Comas. 2016.
\newblock \emph{Database Anonymization: Privacy Models, Data Utility, and
  Microaggregation-based Inter-model Connections}.
\newblock Synthesis Lectures on Information Security, Privacy \& Trust. Morgan
  \& Claypool Publishers.

\bibitem[{Dwork et~al.(2006)Dwork, McSherry, Nissim, and Smith}]{Dwork2006a}
Dwork, Cynthia, Frank McSherry, Kobbi Nissim, and Adam Smith. 2006.
\newblock Calibrating {{Noise}} to {{Sensitivity}} in {{Private Data
  Analysis}}.
\newblock In \emph{Theory of Cryptography}, pages 265--284, Springer Berlin
  Heidelberg, Berlin, Heidelberg.

\bibitem[{Dwork and Roth(2014)}]{10.1561/0400000042}
Dwork, Cynthia and Aaron Roth. 2014.
\newblock The algorithmic foundations of differential privacy.
\newblock \emph{Found. Trends Theor. Comput. Sci.}, 9(3–4):211–407.

\bibitem[{Eder, Krieg-Holz, and Hahn(2020)}]{Ede:Kri:Hah:2020}
Eder, Elisabeth, Ulrike Krieg-Holz, and Udo Hahn. 2020.
\newblock {C}od{E} {A}lltag 2.0 {---} a pseudonymized {G}erman-language email
  corpus.
\newblock In \emph{Proceedings of the 12th Language Resources and Evaluation
  Conference}, pages 4466--4477, European Language Resources Association,
  Marseille, France.

\bibitem[{Elazar and Goldberg(2018)}]{elazar-goldberg-2018-adversarial}
Elazar, Yanai and Yoav Goldberg. 2018.
\newblock Adversarial removal of demographic attributes from text data.
\newblock In \emph{Proceedings of the 2018 Conference on Empirical Methods in
  Natural Language Processing}, pages 11--21, Association for Computational
  Linguistics, Brussels, Belgium.

\bibitem[{Fernandes, Dras, and McIver(2019)}]{DBLP:conf/post/FernandesDM19}
Fernandes, Natasha, Mark Dras, and Annabelle McIver. 2019.
\newblock Generalised differential privacy for text document processing.
\newblock In \emph{Principles of Security and Trust - 8th International
  Conference, {POST} 2019, Held as Part of the European Joint Conferences on
  Theory and Practice of Software, {ETAPS} 2019, Prague, Czech Republic, April
  6-11, 2019, Proceedings}, volume 11426 of \emph{Lecture Notes in Computer
  Science}, pages 123--148, Springer.

\bibitem[{Ferr{\'a}ndez et~al.(2012)Ferr{\'a}ndez, South, Shen, Friedlin,
  Samore, and Meystre}]{Ferrandez}
Ferr{\'a}ndez, O., B.~R. South, S.~Shen, F.~J. Friedlin, M.~H. Samore, and
  S.~M. Meystre. 2012.
\newblock Evaluating current automatic de-identification methods with
  veteran’s health administration clinical documents.
\newblock \emph{BMC Medical Research Methodology}, 12(1):109--124.

\bibitem[{Feyisetan, Diethe, and Drake(2019)}]{feyisetan2019leveraging}
Feyisetan, Oluwaseyi, Tom Diethe, and Thomas Drake. 2019.
\newblock Leveraging hierarchical representations for preserving privacy and
  utility in text.
\newblock In \emph{2019 IEEE International Conference on Data Mining (ICDM)},
  pages 210--219, IEEE.

\bibitem[{Finn, Wright, and Friedewald(2013)}]{finn2013seven}
Finn, Rachel~L, David Wright, and Michael Friedewald. 2013.
\newblock Seven types of privacy.
\newblock In \emph{European data protection: coming of age}. Springer, pages
  3--32.

\bibitem[{GDPR(2016)}]{GDPR}
GDPR. 2016.
\newblock {G}eneral {D}ata {P}rotection {R}egulation.
\newblock European Union Regulation 2016/679.

\bibitem[{Gearty(1993)}]{gearty1993european}
Gearty, Conor~A. 1993.
\newblock The european court of human rights and the protection of civil
  liberties: An overview.
\newblock \emph{The Cambridge Law Journal}, 52(1):89--127.

\bibitem[{Golle(2006)}]{golle2006}
Golle, Philippe. 2006.
\newblock {Revisiting the uniqueness of simple demographics in the US
  population}.
\newblock In \emph{{Proceedings of the 5th {ACM} {W}orkshop on {P}rivacy in
  electronic society}}, pages 77--80, ACM.

\bibitem[{Habernal(2021)}]{Habernal.2021.EMNLP}
Habernal, Ivan. 2021.
\newblock {When differential privacy meets NLP: The devil is in the detail}.
\newblock In \emph{Proceedings of the 2021 Conference on Empirical Methods in
  Natural Language Processing}, page (to appear), Association for Computational
  Linguistics, Punta Cana, Dominican Republic.

\bibitem[{Hart, Manadhata, and Johnson(2011)}]{Hart}
Hart, Michael, Pratyusa Manadhata, and Rob Johnson. 2011.
\newblock {Text Classification for Data Loss Prevention}.
\newblock In \emph{Proceedings of the 11th Privacy Enhancing Technologies
  Symposium (PETS)}, pages 18--37.

\bibitem[{Hassan, S\'anchez, and Domingo-Ferrer(2021)}]{tkde}
Hassan, Fadi, David S\'anchez, and Josep Domingo-Ferrer. 2021.
\newblock Utility-preserving privacy protection of textual documents via word
  embeddings.
\newblock \emph{IEEE Transactions on Knowledge and Data Engineering}, (in
  press).

\bibitem[{Hathurusinghe, Nejadgholi, and Bolic(2021)}]{wikipii}
Hathurusinghe, Rajitha, Isar Nejadgholi, and Miodrag Bolic. 2021.
\newblock A privacy-preserving approach to extraction of personal information
  through automatic annotation and federated learning.

\bibitem[{Hintze(2017)}]{10.1093/idpl/ipx020}
Hintze, Mike. 2017.
\newblock {Viewing the GDPR through a de-identification lens: a tool for
  compliance, clarification, and consistency}.
\newblock \emph{International Data Privacy Law}, 8(1):86--101.

\bibitem[{HIPAA(2004)}]{HIPAA}
HIPAA. 2004.
\newblock \emph{The {H}ealth {I}nsurance {P}ortability and {A}ccountability
  {A}ct}.
\newblock U.S. Dept. of Labor, Employee Benefits Security Administration.

\bibitem[{Honnibal et~al.(2020)Honnibal, Montani, Van~Landeghem, and
  Boyd}]{spacy}
Honnibal, Matthew, Ines Montani, Sofie Van~Landeghem, and Adriane Boyd. 2020.
\newblock {spaCy: Industrial-strength Natural Language Processing in Python}.

\bibitem[{Huang et~al.(2020)Huang, Song, Chen, Li, and
  Arora}]{huang-etal-2020-texthide}
Huang, Yangsibo, Zhao Song, Danqi Chen, Kai Li, and Sanjeev Arora. 2020.
\newblock {T}ext{H}ide: Tackling data privacy in language understanding tasks.
\newblock In \emph{Findings of the Association for Computational Linguistics:
  EMNLP 2020}, pages 1368--1382, Association for Computational Linguistics,
  Online.

\bibitem[{Hundepool et~al.(2012)Hundepool, Domingo-Ferrer, Franconi, Giessing,
  Nordholt, Spicer, and De~Wolf}]{hundepool2012statistical}
Hundepool, Anco, Josep Domingo-Ferrer, Luisa Franconi, Sarah Giessing,
  Eric~Schulte Nordholt, Keith Spicer, and Peter-Paul De~Wolf. 2012.
\newblock \emph{Statistical disclosure control}.
\newblock John Wiley \& Sons.

\bibitem[{Jensen, Zhang, and Plank(2021)}]{Jensen2021DeidentificationOP}
Jensen, Kristian~N{\o}rgaard, Mike Zhang, and Barbara Plank. 2021.
\newblock De-identification of privacy-related entities in job postings.
\newblock In \emph{Proceedings of the 23rd Nordic Conference of Computational
  Linguistics (NODALIDA)}.

\bibitem[{Johnson, Bulgarelli, and Pollard(2020)}]{johnson2020deidentification}
Johnson, Alistair~EW, Lucas Bulgarelli, and Tom~J Pollard. 2020.
\newblock Deidentification of free-text medical records using pre-trained
  bidirectional transformers.
\newblock In \emph{Proceedings of the ACM Conference on Health, Inference, and
  Learning}, pages 214--221.

\bibitem[{Kasper(2007)}]{kasper2007privacy}
Kasper, Debbie~VS. 2007.
\newblock Privacy as a social good.
\newblock \emph{Social thought \& research}, pages 165--189.

\bibitem[{Krishna, Gupta, and Dupuy(2021)}]{krishna-etal-2021-adept}
Krishna, Satyapriya, Rahul Gupta, and Christophe Dupuy. 2021.
\newblock {AD}e{PT}: Auto-encoder based differentially private text
  transformation.
\newblock In \emph{Proceedings of the 16th Conference of the European Chapter
  of the Association for Computational Linguistics: Main Volume}, pages
  2435--2439, Association for Computational Linguistics, Online.

\bibitem[{Lample et~al.(2016)Lample, Ballesteros, Subramanian, Kawakami, and
  Dyer}]{Lam:Bal:Sub:16}
Lample, Guillaume, Miguel Ballesteros, Sandeep Subramanian, Kazuya Kawakami,
  and Chris Dyer. 2016.
\newblock Neural architectures for named entity recognition.
\newblock In \emph{Proceedings of the 2016 Conference of the {N}orth {A}merican
  {C}hapter of the {A}ssociation for {C}omputational {L}inguistics: {H}uman
  {L}anguage {T}echnologies}, pages 260--270, San Diego, California.

\bibitem[{Lee et~al.(2020)Lee, Yoon, Kim, Kim, Kim, So, and
  Kang}]{lee2020biobert}
Lee, Jinhyuk, Wonjin Yoon, Sungdong Kim, Donghyeon Kim, Sunkyu Kim, Chan~Ho So,
  and Jaewoo Kang. 2020.
\newblock Biobert: a pre-trained biomedical language representation model for
  biomedical text mining.
\newblock \emph{Bioinformatics}, 36(4):1234--1240.

\bibitem[{Lee et~al.(2017)Lee, He, Lewis, and Zettlemoyer}]{lee-etal-2017-end}
Lee, Kenton, Luheng He, Mike Lewis, and Luke Zettlemoyer. 2017.
\newblock End-to-end neural coreference resolution.
\newblock In \emph{Proceedings of the 2017 Conference on Empirical Methods in
  Natural Language Processing}, pages 188--197, Association for Computational
  Linguistics, Copenhagen, Denmark.

\bibitem[{Li, Li, and Venkatasubramanian(2007)}]{Li:2007}
Li, Ninghui, Tiancheng Li, and Suresh Venkatasubramanian. 2007.
\newblock t-{C}loseness: {P}rivacy {B}eyond k-{A}nonymity and l-{D}iversity.
\newblock In \emph{23rd International Conference on Data Engineering (ICDE
  2007)}, pages 106--115, IEEE.

\bibitem[{Li et~al.(2021)Li, Tram{\`e}r, Liang, and Hashimoto}]{li2021large}
Li, Xuechen, Florian Tram{\`e}r, Percy Liang, and Tatsunori Hashimoto. 2021.
\newblock Large language models can be strong differentially private learners.
\newblock \emph{arXiv preprint arXiv:2110.05679}.

\bibitem[{Li, Baldwin, and Cohn(2018)}]{li-etal-2018-towards}
Li, Yitong, Timothy Baldwin, and Trevor Cohn. 2018.
\newblock Towards robust and privacy-preserving text representations.
\newblock In \emph{Proceedings of the 56th Annual Meeting of the Association
  for Computational Linguistics (Volume 2: Short Papers)}, pages 25--30,
  Association for Computational Linguistics, Melbourne, Australia.

\bibitem[{Lison et~al.(2021)Lison, Pil{\'a}n, S{\'a}nchez, Batet, and
  Øvrelid}]{acl2021}
Lison, Pierre, Ildik{\'o} Pil{\'a}n, David S{\'a}nchez, Montserrat Batet, and
  Lilja Øvrelid. 2021.
\newblock {Anonymisation Models for Text Data: State of the Art, Challenges and
  Future Directions}.
\newblock In \emph{{Proceedings of the 59th Annual Meeting of the Association
  for Computational Linguistics and the 11th International Joint Conference on
  Natural Language Processing}}.

\bibitem[{Liu et~al.(2019{\natexlab{a}})Liu, Ott, Goyal, Du, Joshi, Chen, Levy,
  Lewis, Zettlemoyer, and Stoyanov}]{DBLP:journals/corr/abs-1907-11692}
Liu, Yinhan, Myle Ott, Naman Goyal, Jingfei Du, Mandar Joshi, Danqi Chen, Omer
  Levy, Mike Lewis, Luke Zettlemoyer, and Veselin Stoyanov. 2019{\natexlab{a}}.
\newblock Roberta: {A} robustly optimized {BERT} pretraining approach.
\newblock \emph{CoRR}, abs/1907.11692.

\bibitem[{Liu et~al.(2019{\natexlab{b}})Liu, Ott, Goyal, Du, Joshi, Chen, Levy,
  Lewis, Zettlemoyer, and Stoyanov}]{liu2019roberta}
Liu, Yinhan, Myle Ott, Naman Goyal, Jingfei Du, Mandar Joshi, Danqi Chen, Omer
  Levy, Mike Lewis, Luke Zettlemoyer, and Veselin Stoyanov. 2019{\natexlab{b}}.
\newblock Roberta: A robustly optimized bert pretraining approach.

\bibitem[{Liu et~al.(2017)Liu, Tang, Wang, and Chen}]{liu2017identification}
Liu, Zengjian, Buzhou Tang, Xiaolong Wang, and Qingcai Chen. 2017.
\newblock De-identification of clinical notes via recurrent neural network and
  conditional random field.
\newblock \emph{Journal of Biomedical Informatics}, 75:S34--S42.

\bibitem[{Loshchilov and Hutter(2019)}]{loshchilov2018decoupled}
Loshchilov, Ilya and Frank Hutter. 2019.
\newblock Decoupled weight decay regularization.
\newblock In \emph{International Conference on Learning Representations}.

\bibitem[{Marimon et~al.(2019)Marimon, Gonzalez-Agirre, Intxaurrondo,
  Rodriguez, Martin, Villegas, and Krallinger}]{Mar:Gon:Int:2019}
Marimon, Montserrat, Aitor Gonzalez-Agirre, Ander Intxaurrondo, Heidy
  Rodriguez, Jose~Lopez Martin, Marta Villegas, and Martin Krallinger. 2019.
\newblock Automatic de-identification of medical texts in spanish: the meddocan
  track, corpus, guidelines, methods and evaluation of results.
\newblock In \emph{IberLEF@ SEPLN}, pages 618--638.

\bibitem[{McMahan et~al.(2017)McMahan, Ramage, Talwar, and
  Zhang}]{mcmahan_learning_2017}
McMahan, H.~Brendan, Daniel Ramage, Kunal Talwar, and Li~Zhang. 2017.
\newblock Learning {{Differentially Private Recurrent Language Models}}.
\newblock \emph{arXiv:1710.06963 [cs]}.

\bibitem[{Medlock(2006)}]{medlock2006introduction}
Medlock, Ben. 2006.
\newblock An introduction to {NLP}-based textual anonymisation.
\newblock In \emph{Proceedings of the Fifth International Conference on
  Language Resources and Evaluation ({LREC}{'}06)}, pages 1051--1056, European
  Language Resources Association (ELRA), Genoa, Italy.

\bibitem[{Megyesi et~al.(2018)Megyesi, Granstedt, Johansson, Prentice,
  Ros{\'e}n, Schenstr{\"o}m, Sundberg, Wir{\'e}n, and
  Volodina}]{megyesi2018learner}
Megyesi, Be{\'a}ta, Lena Granstedt, Sofia Johansson, Julia Prentice, Dan
  Ros{\'e}n, Carl-Johan Schenstr{\"o}m, Gunl{\"o}g Sundberg, Mats Wir{\'e}n,
  and Elena Volodina. 2018.
\newblock Learner corpus anonymization in the age of {GDPR}: Insights from the
  creation of a learner corpus of {S}wedish.
\newblock In \emph{Proceedings of the 7th workshop on {NLP} for Computer
  Assisted Language Learning}, pages 47--56, LiU Electronic Press, Stockholm,
  Sweden.

\bibitem[{Mendels(2020)}]{Mendels}
Mendels, Omri. 2020.
\newblock Custom nlp approaches to data anonymization.
\newblock \emph{Towards data science}.

\bibitem[{Meystre et~al.(2010)Meystre, Friedlin, South, Shen, and
  Samore}]{meystre2010automatic}
Meystre, Stephane~M, F~Jeffrey Friedlin, Brett~R South, Shuying Shen, and
  Matthew~H Samore. 2010.
\newblock Automatic de-identification of textual documents in the electronic
  health record: a review of recent research.
\newblock \emph{BMC Medical Research Methodology}, 10(1):70.

\bibitem[{Mosallanezhad, Beigi, and Liu(2019)}]{mosallanezhad-etal-2019-deep}
Mosallanezhad, Ahmadreza, Ghazaleh Beigi, and Huan Liu. 2019.
\newblock Deep reinforcement learning-based text anonymization against
  private-attribute inference.
\newblock In \emph{Proceedings of the 2019 Conference on Empirical Methods in
  Natural Language Processing and the 9th International Joint Conference on
  Natural Language Processing (EMNLP-IJCNLP)}, pages 2360--2369, Association
  for Computational Linguistics, Hong Kong, China.

\bibitem[{Mozes and Kleinberg(2021)}]{mozes2021no}
Mozes, Maximilian and Bennett Kleinberg. 2021.
\newblock No intruder, no validity: Evaluation criteria for privacy-preserving
  text anonymization.
\newblock \emph{arXiv preprint arXiv:2103.09263}.

\bibitem[{Neamatullah et~al.(2008)Neamatullah, Douglass, Li-wei, Reisner,
  Villarroel, Long, Szolovits, Moody, Mark, and
  Clifford}]{neamatullah2008automated}
Neamatullah, Ishna, Margaret~M Douglass, H~Lehman Li-wei, Andrew Reisner,
  Mauricio Villarroel, William~J Long, Peter Szolovits, George~B Moody, Roger~G
  Mark, and Gari~D Clifford. 2008.
\newblock Automated de-identification of free-text medical records.
\newblock \emph{BMC Medical Informatics and Decision Making}, 8(1):32.

\bibitem[{Papadopoulou et~al.(2022)Papadopoulou, Lison, , Øvrelid, and
  Pilán}]{domainspecific}
Papadopoulou, Anthi, Pierre Lison, , Lilja Øvrelid, and Ildikó Pilán. 2022.
\newblock Bootstrapping text anonymization models with distant supervision.
\newblock In \emph{Proceedings of the Language Resources and Evaluation
  Conference}, pages 4477--4487, European Language Resources Association,
  Marseille, France.

\bibitem[{Patel et~al.(2013)Patel, Accorsi, Inkpen, Lopez, and
  Roche}]{patel2013}
Patel, Namrata, Pierre Accorsi, Diana Inkpen, Cédric Lopez, and Mathieu Roche.
  2013.
\newblock {Approaches of anonymisation of an SMS corpus}.
\newblock In \emph{{Proceedings of the International Conference on Intelligent
  Text Processing and Computational Linguistics}}, pages 77--88.

\bibitem[{Peloquin et~al.(2020)Peloquin, DiMaio, Bierer, and
  Barnes}]{peloquin2020disruptive}
Peloquin, David, Michael DiMaio, Barbara Bierer, and Mark Barnes. 2020.
\newblock Disruptive and avoidable: Gdpr challenges to secondary research uses
  of data.
\newblock \emph{European Journal of Human Genetics}, 28(6):697--705.

\bibitem[{Reddy and Knight(2016)}]{reddy2016obfuscating}
Reddy, Sravana and Kevin Knight. 2016.
\newblock Obfuscating gender in social media writing.
\newblock In \emph{Proceedings of the First Workshop on {NLP} and Computational
  Social Science}, pages 17--26, Association for Computational Linguistics,
  Austin, Texas.

\bibitem[{Resnik(1995)}]{resnik95}
Resnik, Philip. 1995.
\newblock {Using information content to evaluate semantic similarity in a
  taxonomy}.
\newblock In \emph{{Proceedings of the 14th international joint conference on
  Artificial intelligence (IJCAI'95)}}, pages 448--453.

\bibitem[{Rumbold and Pierscionek(2017)}]{rumbold2017effect}
Rumbold, John Mark~Michael and Barbara Pierscionek. 2017.
\newblock The effect of the general data protection regulation on medical
  research.
\newblock \emph{Journal of medical Internet research}, 19(2):e47.

\bibitem[{Samarati(2001)}]{Samarati}
Samarati, Pierangela. 2001.
\newblock Protecting respondents' identities in microdata release.
\newblock \emph{IEEE Transactions on Knowledge and Data Engineering},
  13(6):1010--1027.

\bibitem[{Samarati and Sweeney(1998)}]{Samarati_Sweeney:1998}
Samarati, Pierangela and Latanya Sweeney. 1998.
\newblock {P}rotecting {P}rivacy when {D}isclosing {I}nformation: k-{A}nonymity
  and its {E}nforcement through {G}eneralization and {S}uppression.
\newblock Technical report, SRI International.

\bibitem[{S{\'a}nchez and Batet(2016)}]{Sanchez2016}
S{\'a}nchez, David and Montserrat Batet. 2016.
\newblock C-sanitized: A privacy model for document redaction and sanitization.
\newblock \emph{Journal of the Association for Information Science and
  Technology}, 67(1):148--163.

\bibitem[{S{\'a}nchez and Batet(2017)}]{Sanchez2017}
S{\'a}nchez, David and Montserrat Batet. 2017.
\newblock Toward sensitive document release with privacy guarantees.
\newblock \emph{Engineering Applications of Artificial Intelligence},
  59:23--34.

\bibitem[{S{\'a}nchez, Batet, and Viejo(2013)}]{sanchez2013}
S{\'a}nchez, David, Montserrat Batet, and Alexandre Viejo. 2013.
\newblock Minimizing the disclosure risk of semantic correlations in document
  sanitization.
\newblock \emph{Information Sciences}, 249:110--123.

\bibitem[{Santanen(2019)}]{SANTANEN20195}
Santanen, Eric. 2019.
\newblock The value of protecting privacy.
\newblock \emph{Business Horizons}, 62(1):5--14.

\bibitem[{Shokri et~al.(2017)Shokri, Stronati, Song, and
  Shmatikov}]{shokri2017membership}
Shokri, Reza, Marco Stronati, Congzheng Song, and Vitaly Shmatikov. 2017.
\newblock Membership inference attacks against machine learning models.
\newblock In \emph{2017 IEEE Symposium on Security and Privacy (SP)}, pages
  3--18, IEEE.

\bibitem[{Staddon, Golle, and Zimny(2007)}]{staddon2007}
Staddon, Jessica, Philippe Golle, and Bryce Zimny. 2007.
\newblock {Web-Based Inference Detection}.
\newblock In \emph{{USENIX Security Symposium}}.

\bibitem[{Stubbs, Filannino, and Uzuner(2017)}]{stubbs2017identification}
Stubbs, Amber, Michele Filannino, and {\"O}zlem Uzuner. 2017.
\newblock De-identification of psychiatric intake records: Overview of 2016
  {CEGS N-GRID Shared Tasks Track 1}.
\newblock \emph{Journal of Biomedical Informatics}, 75:S4--S18.

\bibitem[{Stubbs and Uzuner(2015)}]{stubbs2015annotating}
Stubbs, Amber and {\"O}zlem Uzuner. 2015.
\newblock Annotating longitudinal clinical narratives for de-identification:
  The 2014 {i2b2/UTHealth} corpus.
\newblock \emph{Journal of Biomedical Informatics}, 58:S20--S29.

\bibitem[{Sukthanker et~al.(2020)Sukthanker, Poria, Cambria, and
  Thirunavukarasu}]{SUKTHANKER2020139}
Sukthanker, Rhea, Soujanya Poria, Erik Cambria, and Ramkumar Thirunavukarasu.
  2020.
\newblock Anaphora and coreference resolution: A review.
\newblock \emph{Information Fusion}, 59:139--162.

\bibitem[{Sweeney(1996)}]{sweeney1996replacing}
Sweeney, Latanya. 1996.
\newblock Replacing personally-identifying information in medical records, the
  scrub system.
\newblock In \emph{Proceedings of the AMIA annual fall symposium}, pages
  333--337, American Medical Informatics Association.

\bibitem[{Trieu et~al.(2017)Trieu, Tran, Tran, and Tran}]{Trieu}
Trieu, Lap~Q., Trung-Nguyen Tran, Mai-Khiem Tran, and Minh-Triet Tran. 2017.
\newblock {Document Sensitivity Classification for Data Leakage Prevention with
  Twitter-Based Document Embedding and Query Expansion}.
\newblock In \emph{Proceedings of the 13th International Conference on
  Computational Intelligence and Security}, pages 537--542.

\bibitem[{Vartanian and Shabtai(2014)}]{Vartanian}
Vartanian, Arik and Asaf Shabtai. 2014.
\newblock Tm-score: A misuseability weight measure for textual content.
\newblock \emph{IEEE Transactions on Information Forensics and Security},
  9(12):2205--2219.

\bibitem[{Velupillai et~al.(2009)Velupillai, Dalianis, Hassel, and
  Nilsson}]{VELUPILLAI2009e19}
Velupillai, Sumithra, Hercules Dalianis, Martin Hassel, and Gunnar~H. Nilsson.
  2009.
\newblock Developing a standard for de-identifying electronic patient records
  written in {Swedish}: Precision, recall and $f$-measure in a manual and
  computerized annotation trial.
\newblock \emph{International Journal of Medical Informatics}, 78(12):19 -- 26.

\bibitem[{Weischedel et~al.(2011)Weischedel, Hovy, Mitchell, S., Belvin,
  Pradhan, Ramshaw, and Xue}]{ontonotes2011}
Weischedel, Ralph, Eduard Hovy, Marcus. Mitchell, Palmer~Martha S., Robert
  Belvin, Sameer~S. Pradhan, Lance Ramshaw, and Nianwen Xue. 2011.
\newblock {OntoNotes}: A large training corpus for enhanced processing.
\newblock In \emph{Handbook of Natural Language Processing and Machine
  Translation: {DARPA} Global Autonomous Language Exploitation}, Springer.

\bibitem[{Weitzenboeck et~al.(2022)Weitzenboeck, Lison, Cyndecka, and
  Langford}]{weitzenboeck2022gdpr}
Weitzenboeck, Emily, Pierre Lison, Malgorzata~Agnieszka Cyndecka, and Malcolm
  Langford. 2022.
\newblock The {GDPR} and unstructured data: is anonymization possible?
\newblock \emph{International Data Privacy Law}.

\bibitem[{Westin(1967)}]{Westin}
Westin, Alan~F. 1967.
\newblock \emph{Privacy and Freedom}.
\newblock Atheneum, New York.

\bibitem[{Xu et~al.(2019)Xu, Qu, Xu, and Cui}]{xu2019privacy}
Xu, Qiongkai, Lizhen Qu, Chenchen Xu, and Ran Cui. 2019.
\newblock Privacy-aware text rewriting.
\newblock In \emph{Proceedings of the 12th International Conference on Natural
  Language Generation}, pages 247--257, Association for Computational
  Linguistics, Tokyo, Japan.

\bibitem[{Yang and Garibaldi(2015)}]{YANG2015S30}
Yang, Hui and Jonathan~M. Garibaldi. 2015.
\newblock Automatic detection of protected health information from clinic
  narratives.
\newblock \emph{Journal of Biomedical Informatics}, 58:30 -- 38.

\bibitem[{Yogarajan, Mayo, and Pfahringer(2018)}]{Yog:May:Pfa:2018}
Yogarajan, Vithya, Michael Mayo, and Bernhard Pfahringer. 2018.
\newblock A survey of automatic de-identification of longitudinal clinical
  narratives.
\newblock \emph{arXiv preprint arXiv:1810.06765}.

\end{thebibliography}
$\phantom{ }$
\newpage
$\phantom{ }$
\newpage
\appendix

\onecolumn

\section*{Disagreements between annotators}

\begin{figure}[h]
\centering
\includegraphics[width=0.80\textwidth]{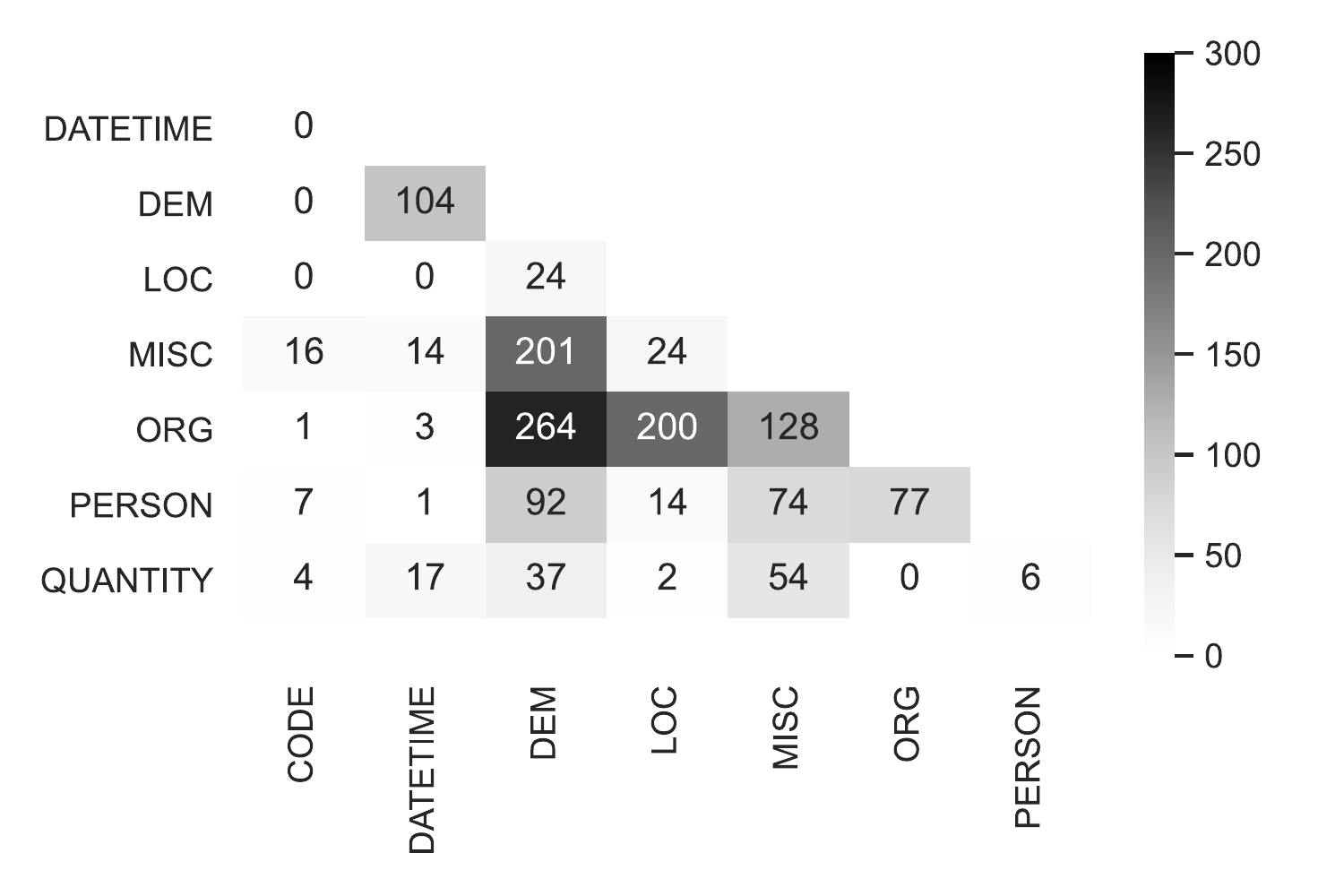}\vspace{-1mm}
\caption{Heatmap for entity type disagreements (number of entity mentions with label mismatch).}\vspace{2mm}
\label{img:heatmap}
\end{figure}

Figure \ref{img:heatmap} shows disagreement on entity types in terms of label mismatch. The most frequent disagreements occurred between \DEM{} and \ORG{}, often when an organization name also reveals demographic information, for example, \textit{naval police, Church of Ireland}. The annotation of the entity \textit{Ombudsman} was found to be particularly challenging, as depending on the context and interpretation, these could fall into not only \DEM{} and \ORG{}, but also \PERSON{}. We can also observe some disagreements between \DEM{} and \MISC{}, especially for entities that did not fall into a clear demographic group (e.g.~\textit{high security prisoner, widower}). Disagreements between \LOC{} and \ORG{}  (\textit{Ireland, Municipality of Göteborg}) are less unexpected due to the inherent ambiguity of those two categories, which have been found to also trigger mixed annotator judgments in NER tasks. 

Table \ref{tbl:masking_disagr_per_entity} presents the number and the proportion of masking disagreements per entity type for entity mentions. We consider in this table only masking disagreements with exact span match. Moreover, the same type of disagreement for the same mention by more than one annotator pair was counted only once. The unique types of disagreement counted this way were in total 4299. Most disagreements were between the \NOMASK{} and \QUASI{} labels for \ORG{}, \DATETIME{}, \DEM{} and \LOC{} entities. 
Approximately 5\% of masking disagreements were between the labels \DIRECT{} and \QUASI{}, mostly for the \PERSON{} and \CODE{} entities.
There were no disagreements between \DIRECT{} and \NOMASK{} labels for identical spans with the same entity label.  

\begin{table}[h]
\caption{\label{tbl:masking_disagr_per_entity} Disagreement on identifier type, factored by entity type. The numbers of parenthesis denote the percentage of mentions marked by two annotators that have this type of disagreement.}
\begin{center}
\begin{tabular}{lrrr}
    \toprule
        Entity type & \DIRECT{} vs. \QUASI{} & \DIRECT{} vs. \NOMASK{} &  \QUASI{} vs. \NOMASK{} \\
        \midrule
        \CODE{} & 65 (1.5) & 0 (0.0) & 70 \ \ (1.6) \\
        \DATETIME{} & 16 (0.4) & 0 (0.0) & 1000 (23.3) \\
        \DEM{} & 0 (0.0) & 0 (0.0) & 497 (11.6) \\
        \LOC{} & 0 (0.0) & 0 (0.0) & 416 \ \ (9.7) \\
        \MISC{} & 0 (0.0) & 0 (0.0) & 158 \ \ (3.7) \\
        \ORG{} & 5 (0.1) & 0 (0.0) & 1068 (24.8) \\
        \PERSON{} & 151 (3.5) & 0 (0.0) & 752 (17.5) \\
        \QUANTITY{} & 0 (0.0) & 0 (0.0) & 101 \ \ (2.4) \\
        \bf Total & 237 \ \ \ \ \ \ \ \ \ & 0 (0.0) & 4062 \ \ \ \ \ \ \ \ \ \  \\
\end{tabular}
\end{center}
\end{table}

\newpage

$\phantom{c}$ \vspace{-14mm}

\section*{Example of document}

$\phantom{c}$ \\[-5mm]

\noindent\textit{After Step 1 (entity detection):}

\begin{small}
\begin{spacing}{2.4}

\noindent\textbf{PROCEDURE}

The case originated in an application (no. \ulnum{CODE}{19840/09}) against the \ulnum{ORG}{United Kingdom of Great Britain and Northern Ireland} lodged with the Court under Article 34 of the Convention for the Protection of Human Rights and Fundamental Freedoms (“the Convention”) by a \ulnum{DEM}{British} national, \ulnum{PERSON}{Mr Harry Shindler} (“the applicant”), on \ulnum{DATETIME}{26 March 2009}.

The applicant was represented by \ulnum{PERSON}{Ms C. Oliver}, a lawyer practising in \ulnum{LOC}{Rome}. The \ulnum{ORG}{United Kingdom Government} (“the Government”) were represented by their Agent, \ulnum{PERSON}{Mr D. Walton}, of the \ulnum{ORG}{Foreign and Commonwealth Office}.

The applicant alleged that his disenfranchisement as a result of his residence outside the \ulnum{ORG}{United Kingdom} constituted a violation of Article 3 of Protocol No. 1 to the Convention, taken alone and taken together with Article 14, and Article 2 of Protocol No. 4 to the Convention.

On \ulnum{DATETIME}{14 December 2010} the application was communicated to the Government. It was also decided to rule on the admissibility and merits of the application at the same time (Article 29 § 1).

$\phantom{}$\\

\noindent\textbf{THE FACTS}

\noindent\textbf{I. THE CIRCUMSTANCES OF THE CASE}

The applicant was born in \ulnum{DATETIME}{1921} and lives in \ulnum{LOC}{Ascoli Piceno, Italy}. He left the \ulnum{ORG}{United Kingdom} in \ulnum{DATETIME}{1982} following his retirement and moved to \ulnum{LOC}{Italy} with his wife, an \ulnum{DEM}{Italian} national.

Pursuant to primary legislation, \ulnum{DEM}{British} citizens residing overseas for less than fifteen years are permitted to vote in parliamentary elections in the \ulnum{LOC}{United Kingdom} (see paragraphs 10-11 below). The applicant does not meet the fifteen-year criterion and is therefore not entitled to vote. In particular, he was unable to vote in the general election of 5 May 2010.

\end{spacing}
\end{small}

$\phantom{c}$ \\[2mm]

\noindent\textit{After Step 2 (masking decisions):}
\begin{small}

\begin{spacing}{2.4}

\noindent\textbf{PROCEDURE}

The case originated in an application (no. \ulnum{DIRECT}{..............} ) against the United Kingdom of Great Britain and Northern Ireland lodged with the Court under Article 34 of the Convention for the Protection of Human Rights and Fundamental Freedoms (“the Convention”) by a \ulnum{QUASI}{..............}  national, \ulnum{DIRECT}{..............}  (“the applicant”), on \ulnum{QUASI}{..............} .

The applicant was represented by  \ulnum{QUASI}{..............} , a lawyer practising in  \ulnum{QUASI}{..............} . The United Kingdom Government (“the Government”) were represented by their Agent,  \ulnum{QUASI}{..............} , of the Foreign and Commonwealth Office.

The applicant alleged that his disenfranchisement as a result of his residence outside the United Kingdom constituted a violation of Article 3 of Protocol No. 1 to the Convention, taken alone and taken together with Article 14, and Article 2 of Protocol No. 4 to the Convention.

On  \ulnum{QUASI}{..............}  the application was communicated to the Government. It was also decided to rule on the admissibility and merits of the application at the same time (Article 29 § 1).

\noindent\textbf{THE FACTS}

\noindent\textbf{I. THE CIRCUMSTANCES OF THE CASE}

The applicant was born in  \ulnum{QUASI}{..............}  and lives in  \ulnum{QUASI}{..............} . He left the United Kingdom in  \ulnum{QUASI}{..............}  following his retirement and moved to  \ulnum{QUASI}{..............}  with his wife, an  \ulnum{QUASI}{..............}  national.

Pursuant to primary legislation, British citizens residing overseas for less than fifteen years are permitted to vote in parliamentary elections in the United Kingdom (see paragraphs 10-11 below). The applicant does not meet the fifteen-year criterion and is therefore not entitled to vote. In particular, he was unable to vote in the general election of 5 May 2010.

\end{spacing}
\end{small}

\newpage
\clearpage
\newpage



\section*{Supplementary material}

# Annotation Guidelines

This text provides annotation guidelines for the anonymisation of human rights court rulings with the Tagtog tool. *Please read the instructions below carefully.*

You are given a collection of court cases (in a directory with your name), each associated with the name of a specific individual to anonymise from the case. The goal of your annotation is to mark all text spans (= a continuous stretch of one or more words) that correspond to an entity of a certain category and, afterwards indicate whether they can be used to re-identify the individual to be protected.

The data has been *pre-annotated* automatically. Your annotation work consists of four stages:

- **Step 0:** Read through the entire case once
- **Step 1: Entities.** Taking the pre-annotations as a starting point, annotate each entity with a semantic type (PERSON, CODE, LOC, ORG, DEM, DATETIME, QUANTITY, MISC, see below). You need to remove/correct errors in the pre-annotated entities, and manually annotate entities that went undetected.
- **Step 2: Masking.** For each entity annotated in Step 1, specify whether their identifier type (direct identifier, quasi-identifier, or does not need masking) and their confidential status. See below for detailed definitions of these terms.
- **Step 3: Final review.** Save and confirm your annotation, and inspect the review document generated in your masked subfolder to ensure you have not missed any identifier or confidential attribute. Otherwise, go back to Step 1 or 2.

## Step 1: Entities

Step 1 focuses on annotating entities with semantic types. To facilitate your task, we provide pre-annotations generated automatically. *Those pre-annotations are just a starting point, and are far from perfect: you <u>must</u> actively seek to correct and extend the annotation*. This means that you should always:

1. Verify whether each pre-annotated entity is correct or needs to be removed or edited (either because it is the wrong type, or whether the span should be changed).
2. Manually annotate each entity that was not detected by the automatic tool. For some categories such as DEM or MISC, the automatic tool will only detect a small fraction of the entities, so you should expect to find some in every document.
3. If two mentions refer to the same underlying entity but have a different string (e.g. "John Smith" and "Mr. Smith"), insert a reference *relation* between them (see below).

For Step 1, you do not need to worry about *re-identification risks* (this will come in Step 2), you just need to mark all entities pertaining to one of the provided categories.

Entities that occur several times through a document are visually marked in TagTog with a yellow border for all mentions except the first one (to indicate that the subsequent mentions are "derived" from the first one). When an entity needs to be corrected or removed, you should

therefore apply your changes on the first mention - this way, your changes will be automatically propagated to all mentions, without having to repeat your corrections one by one.

The list of categories is presented below. If the information does not fit within one of these, the MISC category should be used.

| Category | Description |
| --- | --- |
| PERSON | Names of people, including nicknames/aliases, usernames and initials |
| CODE | Numbers and codes that identify something, such as SSN, phone number, passport number, license plate |
| LOC | Places and locations, such as:<br>• Cities, areas, countries, etc.<br>• Addresses<br>• Named infrastructures (bus stops, bridges, etc.) |
| ORG | Names of organisations, such as:<br>• public and private companies<br>• schools, universities, public institutions, prisons, healthcare institutions non-governmental organisations, churches, etc. |
| DEM | Demographic attribute of a person, such as:<br>• Native language, descent, heritage, ethnicity<br>• Job titles, ranks, education<br>• Physical descriptions, diagnosis, birthmarks, ages |
| DATETIME | Description of a specific date (e.g. October 3, 2018), time (e.g. 9:48 AM) or duration (e.g. 18 years). |
| QUANTITY | Description of a meaninful quantity, e.g. percentages or monetary values. |
| MISC | Every other type of information that describes an individual and that does not belong to the categories above |

## Examples

In general, annotation should mark the *minimal span* which denotes the entity or property in question.

PERSON

For person names the annotation should include titles and honorifics, such as Mr., Dr., etc. since these may contribute to the precise identification of an individual.

- Mr Gestur Jónsson and Mr Ragnar Halldór Hall
- Mr and Mrs Smith

Some examples which are regarded as names:

- Names: E.g., 'Harry Hole', 'Hole', 'Harry'
- Initials: E.g. 'H.H.'
- Spelling mistakes: E.g. 'Hary Hole'
- All orthographic variations E.g. 'harry hole', 'Harry HOLE'

**NB:** The name of a person can be both a direct identifier (if it is the full name of the individual to protect) or a quasi-identifier (if it is the name of another related person).

CODE

Includes ID-numbers and codes, in particular numbers and codes that identify something, such as SSN, phone number, passport number, license plate, report identifiers etc. Codes can be either identifiers or quasi-identifiers, according to whether they unequivocally refer to the individual to protect or not.

- an application (no. 42552/98)

LOC

Includes cities, areas, counties, addresses, as well as other geographical places, buildings and facilities. Other examples are airports, churches, restaurants, hotels, tourist attractions, hospitals, shops, street addresses, roads, oceans, fjords, mountains, parks.

- Reykjavik
- Øvregaten 2a, 5003 Bergen

Include numbers when they are part of the entity name:

- Pilestredet 48(LOC)
- Rema 1000(ORG)

Always annotate the whole name, never nested parts.
Annotate like this:

- Høgskolen i Oslo og Akershus(ORG)

And **not** like this:

- ~~Høgskolen i Oslo(LOC) og Akershus(LOC)~~

Separate entities connected with a conjunction (e.g. 'and') should be annotated separately:

- Pamir and Alay valleys(LOC)

In case an entity could be either ORG or LOC (e.g. Turkey or Breitvet prison), the entity type best describing the referent should be chosen, i.e. ORG if the occurrence refers to the institution itself, and LOC if it refers to a geographic location.

ORG

Includes any named collection of people, businesses, institutions, organizations, universities, hospitals, churches, sports teams, unions, political parties etc.

Corporate designators like AS, Co. and Ltd. are to be included as part of the name.

Translations and acronyms are included in the span, e.g.

- KCK (Koma Civakên Kurdistan – "Kurdistan Communities Union")

Definite or indefinite articles are typically not included in the span, unless explicitly part of the entity (as for titles of books and movies, such as "The Great Gatsby").

- The Supreme Court's Appeals Leave Committee
- A ship from the East India Company
- Istanbul public prosecutor's office
- Istanbul police

DEM

These are demographic properties and include both physical, cultural and occupational/educational properties, such as various physical descriptions, diagnosis, native language, ethnicity, job titles, age, etc.

- 40 years old
- the applicants are journalists
- a group of left-wing extremists
- diagnosed with motor neurone disease
- a Polish and naturalized-French physicist

Pronouns (he, she) should not be annotated to protect gender information.

DATETIME

Prepositions (e.g. on, at) should not be included in the span.

- Monday, October 3, 2018
- at 9:48 AM
- born in 1947

Separate entities connected by "and" should be annotated separately:

- 10 March and 12 of March

Except if they partially overlap:

- 10 and 12 of March

QUANTITY

Units, such as currencies, should be included in the span.

- $37.5 million
- 375 euros
- 4267 SEK
- 1000 Kilos

MISC

Other (quasi-)identifying words such as trademarks, products, events, etc. All things artificially produced are regarded products. This may include more abstract entities, such as speeches, radio shows, programming languages, contracts, laws and even ideas (if they are named).

Brands are products when they refer to a product or a line of products, but organisation when they refer to the acting or producing entity.

- Criminal investigation into the death of the applicant's son

## Exceptions

The main goal of Step 1 is to find all entities related to categories described above (without worrying about whether they need to be masked or not, since this is the goal of Step 2). However, there are some entities that are technically part of those categories, but are so obviously irrelevant to the task that they can be safely ignored.

Here are some concrete exceptions you can ignore:

- profession or title of the legal professionals involved in the case (for instance "solicitor", "legal adviser", "lawyer", etc.)
- parts of generic legal references (such as the year a particular law was passed or published).

If you find other exceptions you think would be useful to add to this list, let us know!

## Relations

Finally, if some text spans are referring to the same underlying entity through different mentions (such as "John Smith" and "Mr Smith", or "Republic of Turkey" and "Turkey"), annotate those referential relations. More precisely:

- a. Find an occurrence of the first mention (such as "John Smith") and click "add relation"
- b. Find an occurrence of the second mention (such as "Mr Smith") and click on it. You should now see a relation between the two.

You do not need to add "same_as" relations between entities with the exact same string. It is also sufficient to annotate the relation once, even though there may be several occurrences of both mentions.

# Step 2: Masking

In this stage of the annotation, you should review all text spans marked in Step 1, and specify whether they need to be masked (either as a DIRECT or QUASI identifier) to protect the identity of the individual specified in the annotation task. You should mark all direct and quasi identifiers but *not more than those* (we still wish to retain as much textual content as possible).

For every entity annotated during stage 1, you should set the correct value for the following two labels:

1. **identifier_type**

    - DIRECT_ID (direct identifiers): text spans that *directly* and *unequivocally* identify the individual to protect) in the case and should therefore be masked.
    - QUASI_ID (quasi-identifiers): text spans that should be masked since they may lead to the re-identification of the individual when combined with other (not masked) quasi-identifiers mentioned in the text along with public background knowledge.
    - NO_MASK: entities that are neither of the above, and should therefore not need to be masked. *Most entities will belong to this category.* Start by applying NO_MASK to the least precise/specific quasi-identifiers in the court case first.

2. **confidential_status**
    If the information is also confidential, that is, it describes religious or philosophical beliefs, political opinions, trade union membership, sexual orientation or sex life, racial or ethnic origin, health data, genetic and biometric data, specify the category (BELIEF, POLITICS, SEX, ETHNIC, HEALTH) or NOT_CONFIDENTIAL if it is not confidential.

Note that the "Confidential Status" label is only available for a subset of entity types for which we can except confidential information.

We recommend you use TagTog's [Document review](#) mode (press r) to easily go through all entities one by one without having to click on anything. When deciding on the identifier type and confidential status of an entity, make sure you select the first mention - this way, your decision will be automatically propagated to all subsequent mentions of that same entity (shown with a yellow border on TagTog).

## Direct identifiers

= text spans that contain information that directly and unequivocally identify the individual to be protected.

**Examples**: person names (including nicknames/aliases and usernames), social security numbers, passport numbers.

## Quasi-identifiers

= Information that, in isolation, does not identify the individual to be protected but can do so, in combination with other quasi-identifiers and background knowledge. These will often refer to demographical ("a 72-years old man") or spatiotemporal attributes ("on February 6 in Sevilla"). For instance, the combination of date of birth, gender and profession will typically allow you to find out the identity of a person.

For a re-identification to be possible, quasi-identifiers must refer to some information that can be seen as potential "publicly available knowledge" — i.e. something that we can expect that an

external person may already know on the individual or may be able to find —, and the combination of quasi-identifying information should be enough to re-identify the individual with no or low ambiguity. You should judge whether it is likely that someone could, based on public knowledge, know the quasi-identifying values of the individual to be protected. There is some room for interpretation here, but the annotator should ask themselves the question: if I wanted to find out the identify of the individual in the document, should I expect to be able to connect those pieces of information with some other knowledge sources (such as news articles, social media, census data, etc.)? and, are those pieces of information enough to re-identify the individual with no or low ambiguity? *In the vast majority of cases, you don't actually need to do any search for those knowledge sources, your intuition will suffice.*

What do we mean by "publicly available knowledge"? In practice, anything that can be found by searching on the web. With one exception: since the texts come from the ECHR, what we consider "publicly available knowledge" should *not* include the HUDOC database itself (or derived knowledge bases) -- otherwise we would need to consider virtually every word in the text as quasi-identifiers, since it would be straightforward to do a quick search on HUDOC to find the original court case and find out the identify of the individual.

As a rule of thumb, immutable personal attributes (e.g., date-of-birth) on an individual that can be known by external entities should be considered quasi-identifiers. Circumstantial attributes may be considered quasi-identifiers or not according to the chance that external entities may know such information (e.g., current place-of-living or a hospital admission date could be, but the number of times one has gone to the grocery store in a week may not).

Usually, only very general attribute values that encompass a large number of individuals (e.g., country-of-birth) may be ignored, since they would match a large population of individuals and would not enable a unequivocal re-identification. This also depends on the presence of other quasi-identifiers within the same document: the larger the amount and the more concrete the information they provide, the larger the chance that they may enable re-identifications.

## Confidential attributes

= Information is such that, if disclosed, could harm or could be a source of discrimination for the individual.

For confidential attributes, the categories of information to be protected are:

| Category | Description |
| --- | --- |
| BELIEF | Religious or philosophical beliefs |
| POLITICS | Political opinions, trade union membership |
| SEX | Sexual orientation or sex life |
| ETHNIC | Racial or ethnic origin |
| HEALTH | Health, genetic and biometric data. This includes sensitive health-related habits, such as substance abuse |

| NOT_CONFIDENTIAL | Not confidential information (most entities) |

## Step 3: Final Review

Save and confirm your annotation. After saving the document, a new "review" document should be now be available in your subfolder masked. For instance, if your document lies in the path annotator1/001-87407, a new document should now appear at annotator1/masked/001-87407_review. (*if that's not the case, let us know*)

This document shows which entity will end up being masked in an anonymised version of the court case. Direct identifiers and quasi-identifiers are now replaced by *****. Confidential attributes are shown in clear text but marked as entity. Inspect the text to ensure you haven't forgotten any (direct or quasi) identifier or confidential attributes. Otherwise, go back to step 1 and 2 and repeat the process.

## Quality check

You will perform a quality check in pairs to review each other's annotations. The procedure to follow is relatively simple: For every *confirmed* document from your teammate, do the following:

1. Read the "review" document carefully
2. Try to answer the question: Is the masking as sufficient to conceal the identity of the person? If in doubt, try to re-identify the person by googling around (but again ignoring the ECHR case itself from what you should consider publicly available knowledge)
3. Go to the "main" document, and check for mistakes/omissions in the annotations. Such mistakes could correspond to missing spans, wrong span boundaries, wrong semantic categories, wrong masking decisions, or wrong/absent confidential status.
4. For "obvious" mistakes or omissions (for which there is no room for multiple interpretations), you can directly edit the annotations to add the corrections
5. For more substantial disagreements where you would have made a different decision than your teammate, note down your disagreement
6. Once you are done reviewing a bulk of court cases, schedule a meeting with your teammate to discuss those disagreements, and reach a final decision by consensus
7. Apply the final edits to the main document
8. Click on confirm on both the main document and the review document (as a way of marking that the review of that document is complete)



\end{document}